\documentclass[10pt]{article}

\usepackage[utf8]{inputenc}
\usepackage{hyperref}
\usepackage{authblk}
\usepackage{setspace}
\usepackage[margin=1.25in]{geometry}
\usepackage{graphicx}
\usepackage{float}
\graphicspath{ {./figures/} }
\usepackage{array}
\usepackage{subcaption}
\usepackage{amsmath}
\usepackage{tabulary}
\usepackage[ruled]{algorithm2e}
\usepackage[usenames,dvipsnames]{color}
\usepackage{amssymb}
\usepackage{bm}
\usepackage{tikz}
\usetikzlibrary{shapes, arrows, positioning, calc}
\usepackage[numbers,sort&compress]{natbib}

\usepackage{booktabs}
\usepackage[labelfont=bf,textfont=it]{caption}
\usepackage{placeins}

\usepackage{hyperref}
\hypersetup{
    colorlinks,
    linkcolor={magenta},
    citecolor={blue},
    urlcolor={blue!80!black},
    breaklinks=true,
    plainpages=false
}
\newcommand{\cref}[3]{\hyperref[#2]{#1~\ref*{#2}{#3}}}
\newcommand{\crefs}[3]{\hyperref[#2]{#1~\ref*{#2}-\ref*{#3}}}
\newcommand{\colref}[2]{\hyperref[#2]{#1~\ref*{#2}}}

\newcommand{\figref}[1]{\colref{Figure}{#1}}

\newcommand{\secref}[1]{\colref{Section}{#1}}

\newcommand{\tabref}[1]{\colref{Table}{#1}}

\newcommand{\doi}[1]{\textsc{doi}: \href{http://dx.doi.org/#1}{\nolinkurl{#1}}}

\title{
MaizeField3D: A Curated 3D Point Cloud and Procedural Model Dataset of Field-Grown Maize from a Diversity Panel
}

\author[1$\dag$]{\small Elvis Kimara}
\author[3$\dag$]{\small Mozhgan Hadadi}
\author[3]{\small Jackson Godbersen}
\author[2]{\small  Aditya Balu}
\author[2]{\small Talukder Jubery}
\author[4,5,6]{\small Yawei Li}
\author[2,3,4]{\\ \small Adarsh Krishnamurthy}
\author[4,5,6]{\small Patrick S. Schnable}
\author[2,3,4,*]{\small Baskar Ganapathysubramanian}

\affil[1]{\small Department of Computer Science, Iowa State University, Ames, USA}
\affil[2]{\small Translational AI Research and Education Center, Ames, USA}
\affil[3]{\small Department of Mechanical Engineering, Iowa State University, Ames, USA}
\affil[4]{\small Plant Science Institute, Iowa State University, Ames, USA}
\affil[5]{\small Interdepartmental Genetics and Genomics Graduate Program, Iowa State University, Ames, USA}
\affil[6]{\small Department of Agronomy, Iowa State University, Ames, USA}

\affil[$\dag$]{Equal Contribution}
\affil[*]{Corresponding Author}

\date{}

\begin{document}
\maketitle
\begin{abstract}

The development of artificial intelligence (AI) and machine learning (ML) based tools for 3D phenotyping, especially for maize, has been limited due to the lack of large and diverse 3D datasets. 2D image datasets fail to capture essential structural details such as leaf architecture, plant volume, and spatial arrangements that 3D data provide. To address this limitation, we present MaizeField3D (\href{https://baskargroup.github.io/MaizeField3D/}{website}), a curated dataset of 3D point clouds of field-grown maize plants from a diverse genetic panel, designed to be AI-ready for advancing agricultural research. Our dataset includes 1,045 high-quality point clouds of field-grown maize collected using a terrestrial laser scanner (TLS). Point clouds of 520 plants from this dataset were segmented and annotated using a graph-based segmentation method to isolate individual leaves and stalks, ensuring consistent labeling across all samples. This labeled data was then used for fitting procedural models that provide a structured parametric representation of the maize plants. The leaves of the maize plants in the procedural models are represented using Non-Uniform Rational B-Spline (NURBS) surfaces that were generated using a two-step optimization process combining gradient-free and gradient-based methods. We conducted rigorous manual quality control on all datasets, correcting errors in segmentation, ensuring accurate leaf ordering, and validating metadata annotations. The dataset also includes metadata detailing plant morphology and quality, alongside multi-resolution subsampled point cloud data (100k, 50k, 10k points), which can be readily used for different downstream computational tasks. MaizeField3D will serve as a comprehensive foundational dataset for AI-driven phenotyping, plant structural analysis, and 3D applications in agricultural research.

\end{abstract}

\section{Introduction}
Understanding phenotypic traits is essential for improving crop yield, resilience, and sustainability in agricultural research~\citep{kusmec2018harnessing, blancon2024maize}. Phenotypic analysis facilitates the study of key characteristics such as plant architecture, leaf angles, and biomass allocation, which are directly linked to yield and environmental adaptability~\citep{grys2017machine, westhues2021prediction, corcoran2023automated, gupta2024ai, tucker2020evaluating}. Among staple crops, maize occupies a pivotal role due to its economic importance, global cultivation, and significant contribution to food security~\citep{westhues2021prediction, tucker2020evaluating}. With the increasing global demand for sustainable agricultural practices, robust phenotyping methods are indispensable for advancing crop improvement programs.

The emergence of advanced 3D imaging technologies, including terrestrial LASER scanning (TLS)~\citep{medic2023challenges} and photogrammetry~\citep{scholz2019determination}, has revolutionized plant phenotyping by offering detailed structural insights beyond the limitations of traditional image-based 2D methods~\citep{bolkas2021comparison, young2023canopy, young2024soybean}. 3D phenotyping enables the precise characterization of complex plant traits, such as canopy structure, volumetric growth, and organ-specific morphology. These methods have been successfully applied to diverse crops, including rubber tree forests, weeds, wheat, and rice, demonstrating their versatility in agricultural research~\citep{lei20233d, dobbs2022new, hu2020nondestructive, friedli2016terrestrial}. By capturing spatial and volumetric features with high accuracy, 3D phenotyping enables detailed plant trait analysis.

Despite these advances, 3D phenotyping poses several challenges, including data complexity, variability in plant structures, and the computational demands of processing high-resolution datasets. Machine learning (ML) and artificial intelligence (AI) have emerged as transformative tools to address these challenges, enabling automated extraction of meaningful phenotypic traits from large and complex 3D datasets~\citep{cembrowska2023integrated, luo2023eff}. AI-powered methods have been employed for tasks such as predicting biomass, analyzing leaf traits, and detecting early stress responses~\citep{gupta2024ai}. Innovative approaches, including weakly supervised learning frameworks such as Eff-3DPSeg~\citep{luo2023eff} and deep learning models for point cloud analysis~\citep{zhang2023three}, have further advanced organ-level segmentation and trait quantification. These technologies significantly enhance the efficiency and accuracy of 3D phenotyping but are hindered by the lack of high-quality, annotated datasets.

A key bottleneck in leveraging AI for 3D plant phenotyping is the scarcity of extensive and agronomically contextual datasets~\citep{Zhu2024, mostafa2023explainable}. Effective training and evaluation of ML models require datasets that capture the complexity of plant structures, span various growth stages, and include detailed annotations such as semantic and instance labels. Well-annotated datasets also facilitate benchmarking and development of algorithms tailored to agricultural applications, bridging the gap between theoretical advancements and real-world deployment.

In the past few years, there has been increasing interest in creating 3D plant datasets to address phenotyping needs. However, creating comprehensive real-world 3D datasets presents significant challenges, especially in agriculture. The process requires substantial resources, including field and greenhouse space, expert agricultural knowledge, and extended periods for plant growth cycles, making it both expensive and complex. To overcome these limitations, researchers have explored synthetic alternatives that can generate large-scale datasets efficiently. 

One approach involves generating simulation-ready 3D plant datasets and systematically constructing CAD models to populate virtual fields. Using tools like SolidWorks and Blender, these models can be embedded into configurable simulation environments (e.g., Gazebo), facilitating visual navigation and field mapping experiments for autonomous phenotyping platforms. Such environments allow flexible modeling of real-world field conditions with configurable crop and weed types, layouts, and occlusions using mesh libraries and randomized pose distributions ~\citep{li5093760visual}. Although mesh or CAD-based synthetic datasets are scalable and customizable, they often lack the realism and variability found in actual agricultural scenes. To improve the generalizability of models trained on synthetic data, researchers have increasingly adopted domain adaptation strategies such as Real2Sim ~\citep{james2019sim}, where real-world measurements guide synthetic model generation. For instance, ~\citep{qiu2024real2sim} proposed a closed-loop Real2Sim framework that uses apple tree point clouds to generate structurally accurate 3D synthetic trees, enabling improved trait extraction under zero-shot conditions without additional real-world data. A recent and promising direction leverages diffusion-based generative models to create visually realistic plant scenes from textual or structured inputs. For instance, \citet{anagnostopoulou2023realistic} developed a diffusion-based framework to synthesize 15,000 photorealistic images of mushrooms, capturing a wide range of natural variability in lighting, background, and shape. 

Complementing these efforts, procedural generation techniques offer a flexible means to synthesize complex, variable plant geometries at scale. These methods use rule-based or stochastic modeling approaches to generate diverse plant morphologies and environments without the need for dense real-world scans. For example, \citet{zarei2024plantsegnet} utilized procedural modeling to generate 64,000 annotated sorghum plants. More recently, \citet{liu2025neural} introduced a neural hierarchical decomposition framework that reconstructs biologically plausible 3D plant structures from single 2D images. By learning to infer a tree-structured graph of organs and their geometric properties, the method enables high-fidelity, simulation-ready plant models from minimal inputs, offering a scalable approach for dataset augmentation and trait analysis.

Our work follows a similar direction. Rather than generating plants entirely from scratch, we reconstruct plant geometry using procedural NURBS surfaces derived from scans of the real field-grown plants. This allows us to maintain biological realism while also enabling scalable generation of synthetic data for tasks like canopy-level light simulation, phenotype extraction, and ML applications. By rooting the synthetic models in real-world structure, we aim to offer a middle ground between pure simulation and costly manual annotation.

\begin{table}[t!]
\centering
\footnotesize
\setlength{\extrarowheight}{3pt}
\caption{Comparison of publicly available 3D plant datasets with annotated 3D models.}
\label{tab:comparison_datasets}
\begin{tabulary}{1.0\linewidth}{p{0.13\linewidth}>{\raggedright\arraybackslash}p{0.18\linewidth}>{\raggedright\arraybackslash}p{0.11\linewidth}p{0.06\linewidth}p{0.38\linewidth}}
\toprule
\textbf{Dataset} & \textbf{\#\,Models (species)}  &  \textbf{Modality} & \textbf{Color} & \textbf{Labels} \\
\midrule
Pheno4D          & 84~(Maize), 140~(Tomato)         & Laser  & No  & Semantic \& Instance. Includes 49 labeled maize point clouds and 77 labeled tomato point clouds. \\
\hline
ROSE-X           & 11~(Rose)                         & X-rays & No  & Semantic. Voxel-level annotations with organ-level labels (stem, leaf, flower). \\
\hline
Single Maize Point Cloud & 428~(Maize)               & Laser  & No  & Semantic \& Instance. Includes semantic and instance annotations of stem and leaves. \\ 
\hline
Plant3D           & 152~(Arabidopsis), 311~(Tomato), 105~(Tobacco), 141~(Sorghum) & Laser  & No  & Growth conditions. Scanned across 20–30 developmental time points under conditions like ambient light, high heat, high light, vegetative shade, and drought.     \\
\hline
Soybean-MVS       & 102~(Soybean)                    & MVS    & Yes & Semantic. Includes 5 varieties, with semantic labels for plant organs like leaves, main stems,  and stems.          \\
\hline
PLANesT-3D        & 10~(Pepper), 10~(Rose), 14~(Ribes) & SfM-MVS& Yes & Semantic \& Instance. Annotated 3D color point clouds with labels for "leaf" and "stem," and uniquely labeled individual leaflets. \\
\hline
Crops3D           & 83~(Tomato), 196~(Cabbage), 118~(Potato), 20~(Potato~plot), 176~(Cotton), 150~(Rapeseed), 14~(Rapeseed plot), 148~(Wheat), 225~(Maize), 16~(Maize plot), 84~(Rice),  & SfM-MVS, Structured-Light, TLS   & Yes & Semantic \& Instance. Includes organ-level segmentation for soil, stem, leaf, tassel, fruit, pot, and more. Instance annotations available for individual plants within plots.\\
\hline
\textbf{MaizeField3D (Ours)} &1,045 (Maize) & TLS, Procedural models    & Yes & Semantic \& Instance. Includes 520 annotated point clouds with segmentation for leaves and stalks. Provides multiple subsampled datasets (100k, 50k, 10k) tailored for different downstream applications. High-resolution AI-ready data ideal for phenotypic analysis and ML applications.\\
\bottomrule
\end{tabulary}
\end{table}

\tabref{tab:comparison_datasets} provides a comparison of existing publicly available 3D plant datasets, highlighting their key characteristics and limitations. These datasets vary widely in size, scanning technology, and annotation detail. The smallest dataset, ROSE-X, includes 11 rose plant models, while Crops3D is the largest, with 1,230 3D plant point clouds across eight crop species. Among single-crop datasets, the Single Maize Point Cloud dataset, with 428 maize samples, represents the largest publicly available collection dedicated to a single species. Pheno4D~\citep{schunck2021pheno4d} introduces a dataset featuring 7 maize plants and 7 tomato plants. These plants were scanned daily using a 3D laser scanning system over two weeks for maize and three weeks for tomato plants, resulting in 84 point clouds for maize (with 49 labeled) and 140 point clouds for tomato (with 77 labeled). The dataset does not include color. Labels applied to the dataset include semantic and instance labels. ROSE-X~\citep{dutagaci2020rose} provides a dataset of 11 complete 3D models of real rosebush plants. These models were acquired using X-ray computed tomography (CT), offering voxel-level annotations for the entire plant shoot, with organ-level labels for stem, leaf, and flower. While the organ labels are color-coded in the point cloud representations, the dataset does not specifically include true color information. A dataset of 428 maize samples was introduced by~\citep{yang2024maize}, where point clouds were collected using a 3D laser scanner. These samples spanned five maize varieties (Xian Yu 335, LD145, LD502, LD586, and LD1281) and were annotated using the Label3DMaize toolkit for semantic and instance segmentation. Downsampled point clouds were processed to a uniform size of 20,480 points per sample, ensuring efficient use in training. A total of 709 plant shoot architectures from four species  (Arabidopsis, Tomato, Tobacco, and Sorghum) are featured in Plant3D~\citep{conn2017high, conn2017statistical, conn2019network}. These architectures have been scanned across 20–30 developmental time points and under a variety of situations, including ambient light, high heat, high light, vegetative shade, and drought.  Soybean-MVS~\citep{sun2023soybean} introduces a dataset consisting of 102 soybean 3D plant models reconstructed using multiple-view stereo (MVS) technology. The dataset spans the entire soybean growth period and includes five varieties. Color data was included in the point clouds (xyzRGB format). The dataset features semantic labels for plant organs, including leaves, main stems, and stems. A dataset of 34 actual plant point clouds from three species—Capsicum annuum (pepper), Rosa kordana (rose), and Ribes rubrum (ribes)—is presented by PLANesT-3D~\citep{mertouglu2024planest}. These plants were reconstructed using Structure from Motion (SfM) and MVS techniques, offering annotated 3D color point clouds. Points were labeled as either "leaf" or "stem", and individual leaflets within the plants were uniquely labeled. Crops3D~\citep{zhu2024crops3d} provides a comprehensive dataset comprising 1,230 colored 3D crop point cloud samples from eight crop species, including cabbage, cotton, maize, potato, rapeseed, rice, tomato, and wheat. Point clouds were acquired using a combination of TLS, Structured Light Scanning, and SfM-MVS techniques. Points were categorized into specific plant organs such as soil, stem, leaf, tassel, fruit, pot, and others, depending on the crop. The dataset includes instance annotations for individual plants within plots.

While these datasets have advanced plant-specific ML algorithms and segmentation methods, they are often limited by challenges such as low-resolution models, incomplete annotations, small dataset sizes, and restricted diversity of plant conditions. For maize, existing datasets often lack comprehensive annotations or fail to incorporate field-grown plants representing real-world phenotypic variability. These gaps hinder the development of robust phenotyping solutions, particularly for traits like organ-level segmentation and trait analysis. 

To overcome these limitations, we introduce MaizeField3D, a curated, large-scale dataset with 1,045 high-resolution 3D point clouds of field-grown maize plants captured using TLS and additionally represented using procedural modeling techniques. This dataset was collected from the SAM diversity panel. It includes 520 annotated point clouds with semantic and instance segmentation for leaves and stalks. Each segmented point cloud includes RGB color labels that encode leaf identity based on vertical position, facilitating organ-level analysis and visualization. Additionally, MaizeField3D incorporates procedurally generated maize models, which were created for each of the 520 segmented maize plants as described in \secref{procedural}. This procedural modeling approach allows for low dimensional, surface based representation of realistic maize canopies by leveraging actual plant data. The procedural models provide advantages for ML approaches, including data augmentation and synthetic-to-real transfer learning. To further support data-driven research, MaizeField3D provides multiple subsampled versions (100k, 50k, and 10k points per cloud) optimized for different computational needs.

The main contributions of MaizeField3D are:
\begin{itemize}
    \item A curated dataset of 1,045 high-quality 3D point clouds representing diverse, field-grown maize plants, with 520 of them annotated at the organ level with detailed semantic and instance segmentation for leaves and stalks.
    \item Procedurally generated NURBS surface models of 520 maize plants' leaves, offering a compact, structured, and biologically meaningful representation for downstream applications such as functional-structural analysis and detailed (bio)physics simulation.
    \item RGB color labels encoding leaf identity based on vertical position, supporting consistent organ-level visualization and analysis.
    \item Standardized data formats and multiple resolution levels (100k, 50k, and 10k points per cloud), enabling flexible integration into machine learning pipelines with varying computational requirements.
    \item A demonstrated use case involving canopy-scale radiation interception simulation using HELIOS~\citep{bailey2019helios}, where virtual field plots constructed from our procedural models are used to compute absorbed photosynthetically active radiation (PAR) under real solar conditions.
\end{itemize}

In the following sections, we explain the methods for data processing, dataset annotation, metadata creation, and quality control. By providing this dataset, we aim to catalyze the development of innovative phenotyping solutions, paving the way for new insights into maize biology and crop improvement.

\section{Dataset Description}

\subsection{Dataset Access}

The MaizeField3D dataset\footnote{\url{https://baskargroup.github.io/MaizeField3D/}} is publicly available through the Hugging Face Datasets platform\footnote{\url{https://huggingface.co/datasets/BGLab/MaizeField3D}}, ensuring accessibility and ease of integration into ML pipelines. The dataset includes high-resolution 3D point clouds and segmented plant models, organized for efficient navigation and usability. Users can download the complete dataset or select specific components tailored to their research needs. Key features include:

\begin{itemize}
    \item \textbf{Point Clouds}: Point cloud data of 1,045 maize plants at multiple subsampled resolutions (\(100,000\), \(50,000\), and \(10,000\) points) to accommodate various computational requirements.
    \item \textbf{Segmentation}: Detailed segmented models of 520 plants, including individual leaves and stalks, with subsampled resolutions (\(100,000\), \(50,000\), and \(10,000\) points). Each leaf is consistently color-coded based on its vertical position, and stalks are assigned distinct colors, facilitating organ-specific visualization and analysis.
    \item \textbf{Metadata}: A comprehensive metadata file (\texttt{Metadata.xlsx}) provides detailed annotations for each plant, ensuring clarity and contextual understanding of the dataset.
    \item \textbf{Documentation}: A detailed user guide is included in the repository, offering step-by-step instructions for integrating the dataset with tools such as PyTorch, TensorFlow, and Scikit-learn, enabling streamlined ML applications.
\end{itemize}

To enhance the dataset's usability, the following reconstructed outputs of integration of the segmented dataset with the procedural model explained in \secref{procedural} are also included:
\begin{itemize}
    \item \textbf{STL Files:} These files include the 3D reconstructed surfaces, compatible with visualization and simulation tools.
    \item \textbf{DAT Files:} These files provide the control point information for the NURBS surfaces, enabling further refinements and integration into CAD workflows.
\end{itemize}

The Hugging Face repository supports both full dataset downloads and selective access to subsets, such as segmented plants or point clouds at specific resolutions. This flexibility allows users to tailor the dataset to their specific requirements. For any inquiries or assistance in accessing specific components, users are encouraged to contact the corresponding authors via the email addresses provided in this publication.

\begin{figure}[b!]
    \centering
    \begin{subfigure}[b]{\textwidth}
        \centering
        \begin{subfigure}[b]{0.3\textwidth}
            \includegraphics[trim=10in 0in 10in 0in, clip, width=\textwidth]{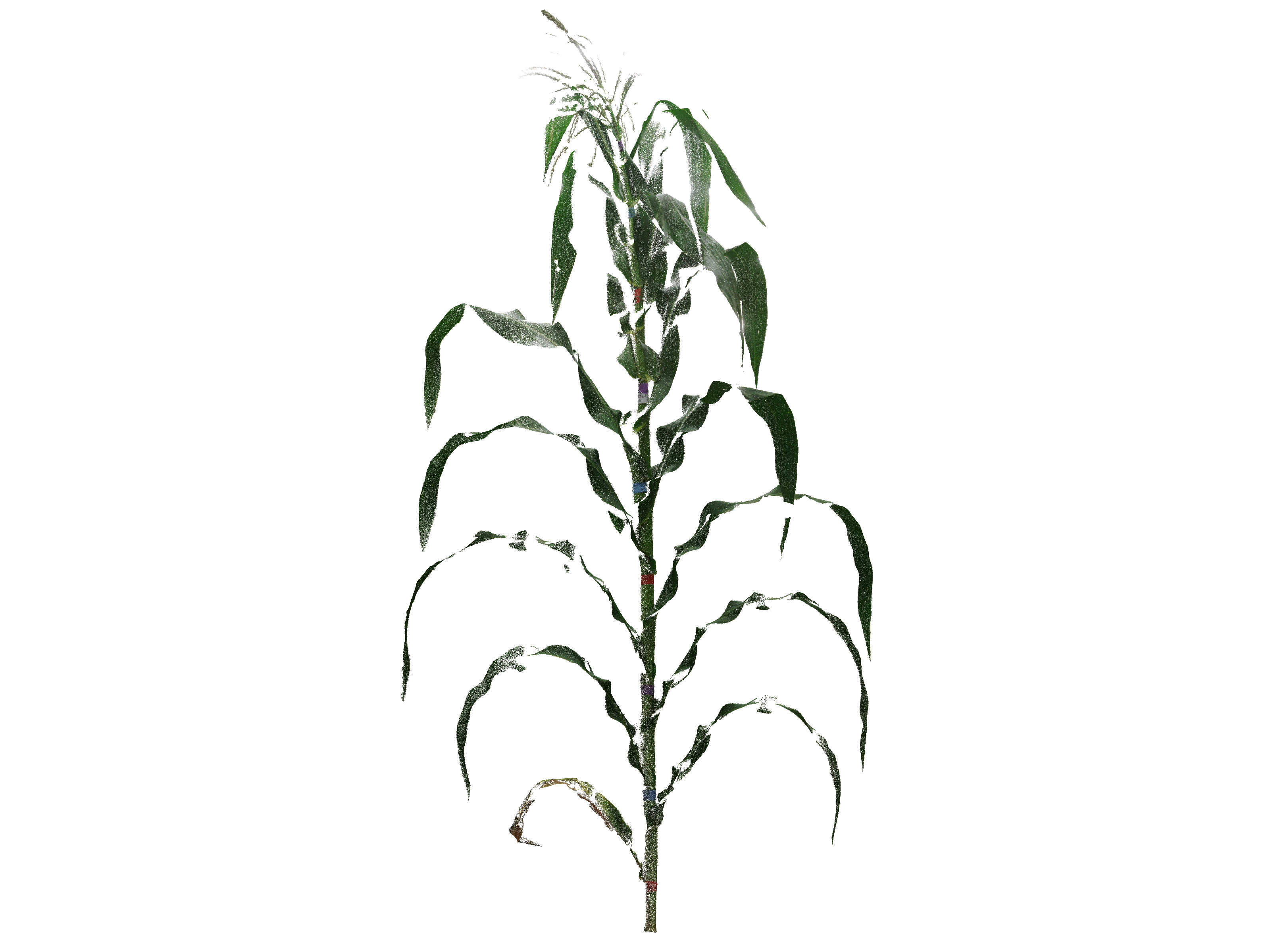}
        \end{subfigure}
        \begin{subfigure}[b]{0.3\textwidth}
            \includegraphics[trim=10in 0in 10in 0in, clip, width=\textwidth]{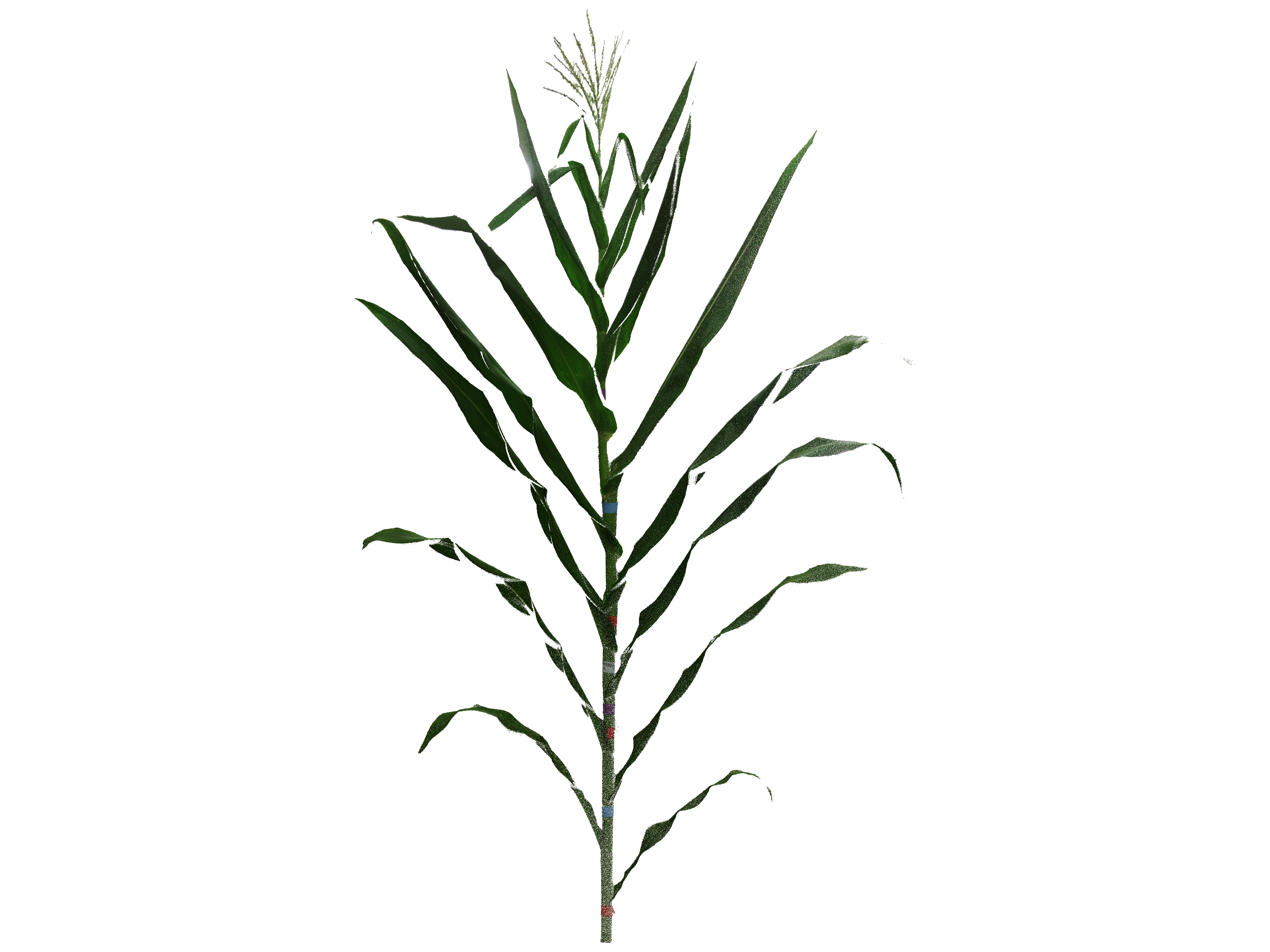}
        \end{subfigure}
        \begin{subfigure}[b]{0.3\textwidth}
            \includegraphics[trim=10in 0in 10in 0in, clip, width=\textwidth]{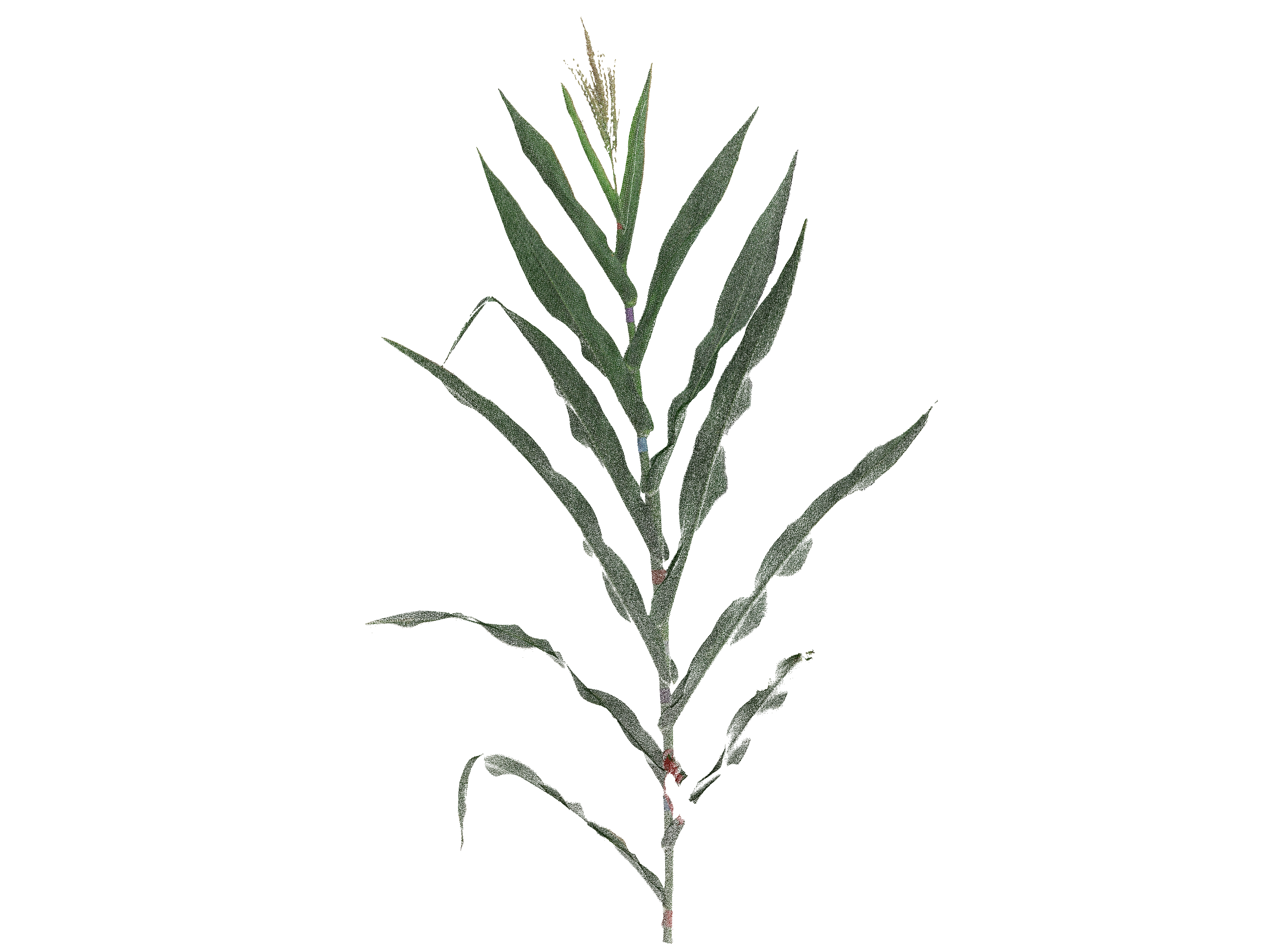}
        \end{subfigure}
        \caption{}
        \label{fig:original_plants}
    \end{subfigure}
    \vspace{0.8em}
    \begin{subfigure}[b]{\textwidth}
        \centering
        \begin{subfigure}[b]{0.3\textwidth}
            \includegraphics[trim=10in 0in 10in 0in, clip, width=\textwidth]{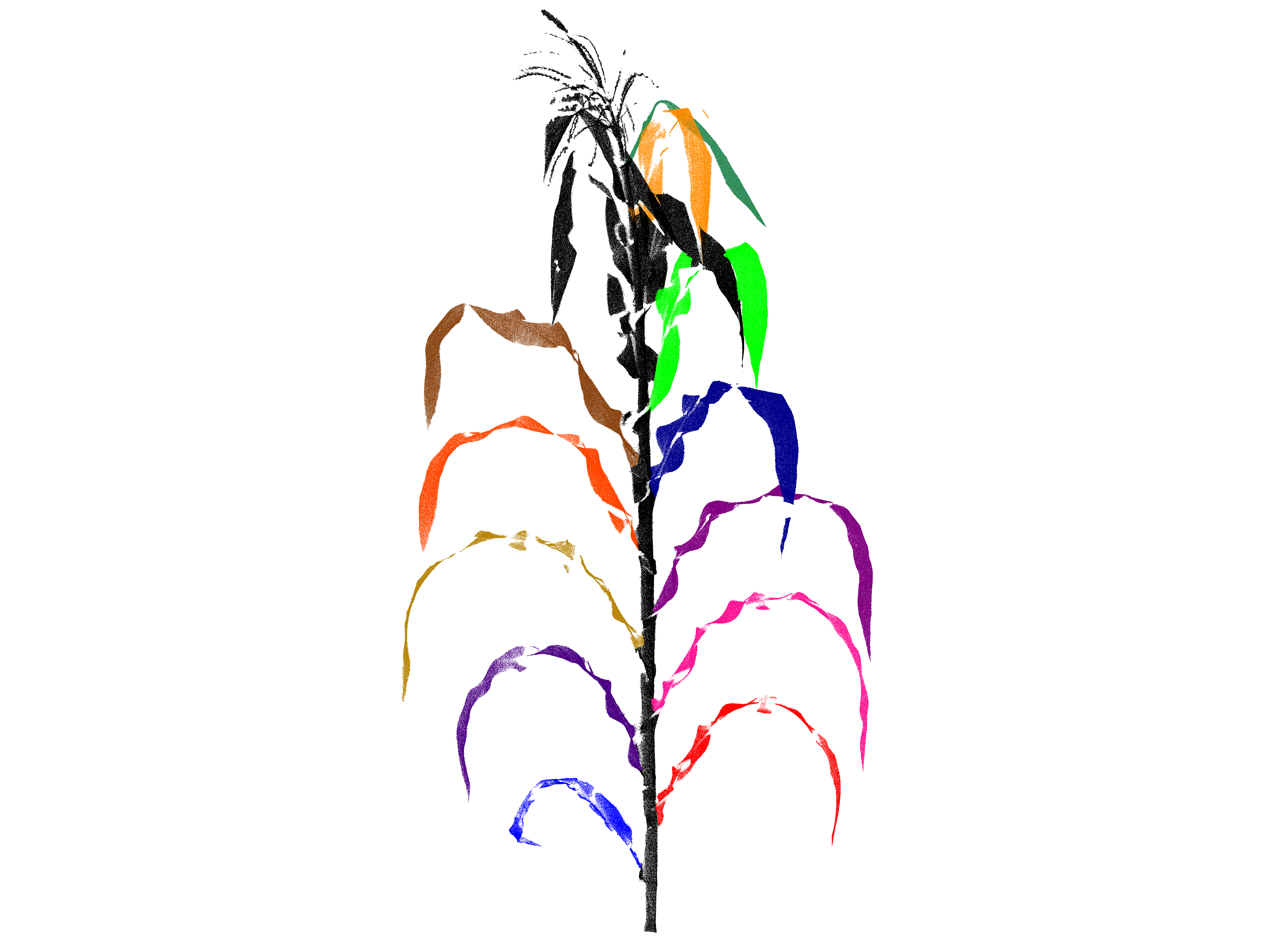}
        \end{subfigure}
        \begin{subfigure}[b]{0.3\textwidth}
            \includegraphics[trim=10in 0in 10in 0in, clip, width=\textwidth]{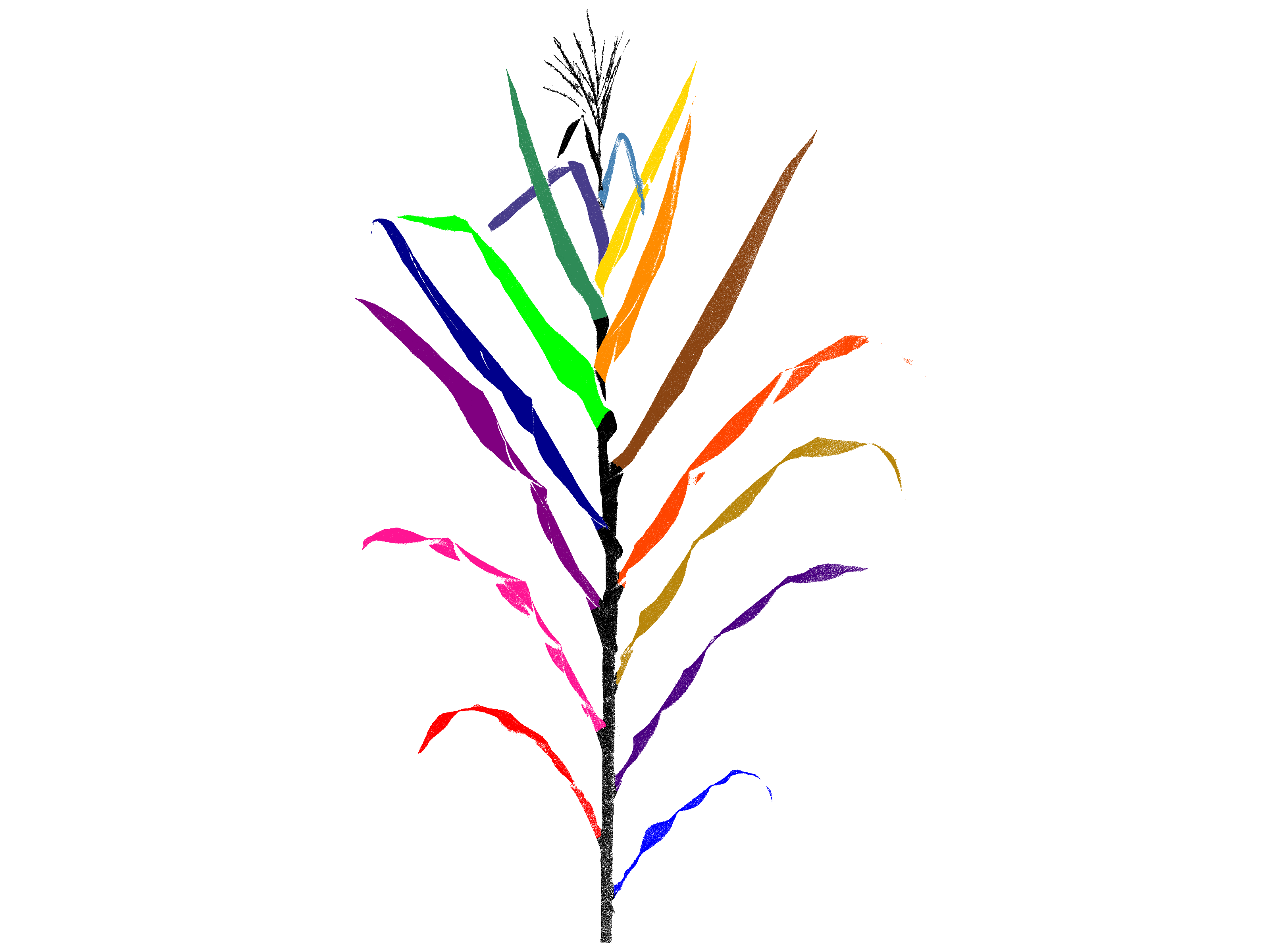}
        \end{subfigure}
        \begin{subfigure}[b]{0.3\textwidth}
            \includegraphics[trim=10in 0in 10in 0in, clip, width=\textwidth]{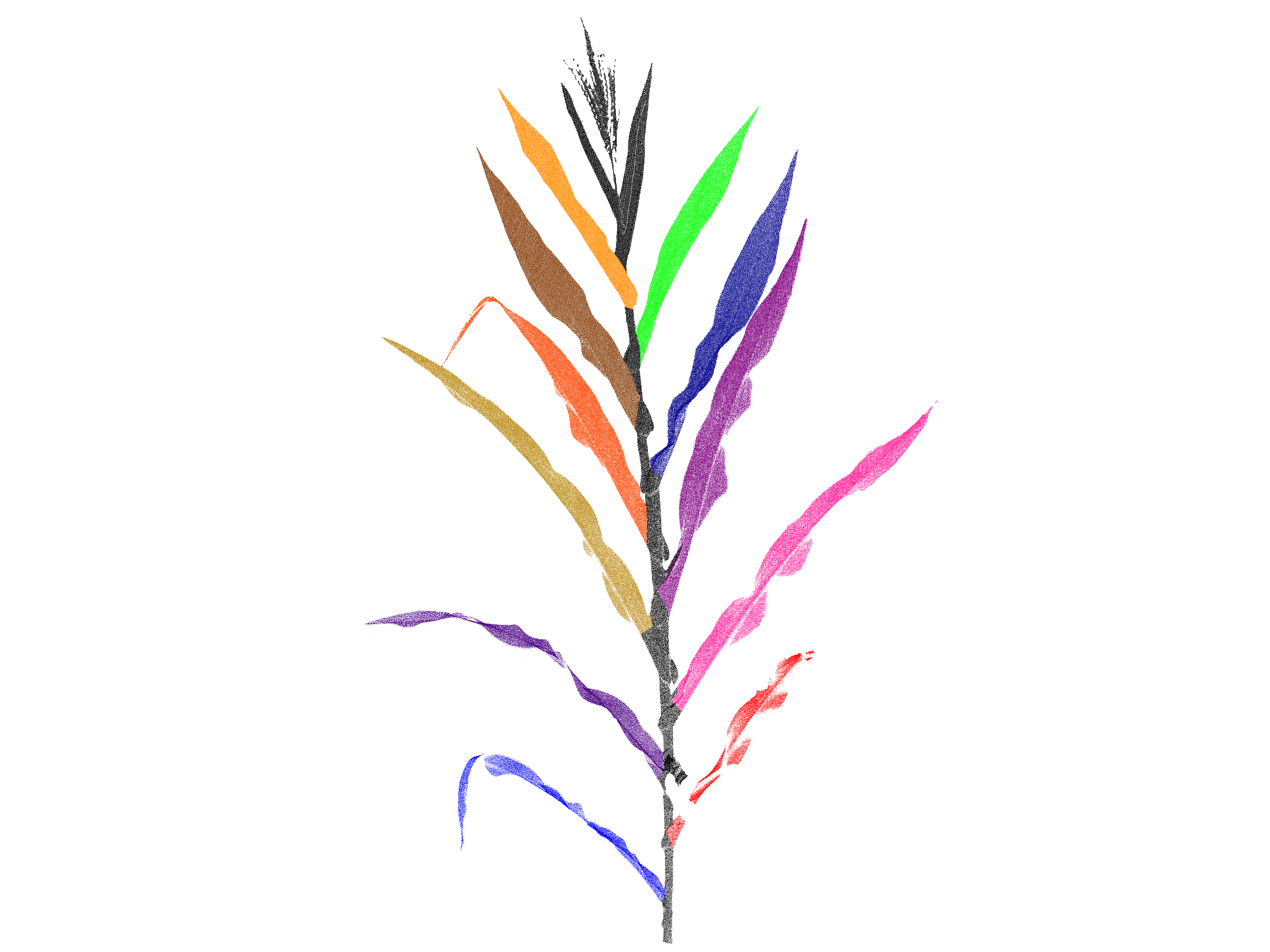}
        \end{subfigure}
        \caption{}
        \label{fig:segmented_plants}
    \end{subfigure}
    \caption{(a) Original point clouds of three maize plants, highlighting the diverse morphologies of maize plants captured using terrestrial laser scanning. (b) Corresponding segmented maize plant point clouds, showcasing the color-coded segmentation data.}
    \label{fig:combined_original_segmented}
\end{figure}

\subsection{Dataset Composition}
The dataset comprises 1,045 high-resolution point clouds, each capturing detailed 3D information of individual maize plants. The dataset represents a diverse range of genetic backgrounds and provides the following key components:
\begin{itemize}
    \item High-resolution point clouds for various maize plants. 
    \item 520 segmented plants into stalks and leaves with specific color coding, ordered from bottom to top leaves.
    \item Metadata detailing plant quality, presence of tassels and ears, and other critical attributes.
    \item Multiple resolutions to support diverse applications.
\end{itemize}
\figref{fig:original_plants} showcases three example maize plants from the dataset, highlighting the original data without segmentation and the diversity in plant morphology. Additional examples are provided in \figref{fig:additional_original_plants} (Supplementary Materials).

\begin{figure}[t!]
    \centering
    \begin{subfigure}[b]{0.3\textwidth}
        \centering
        \includegraphics[width=\textwidth]{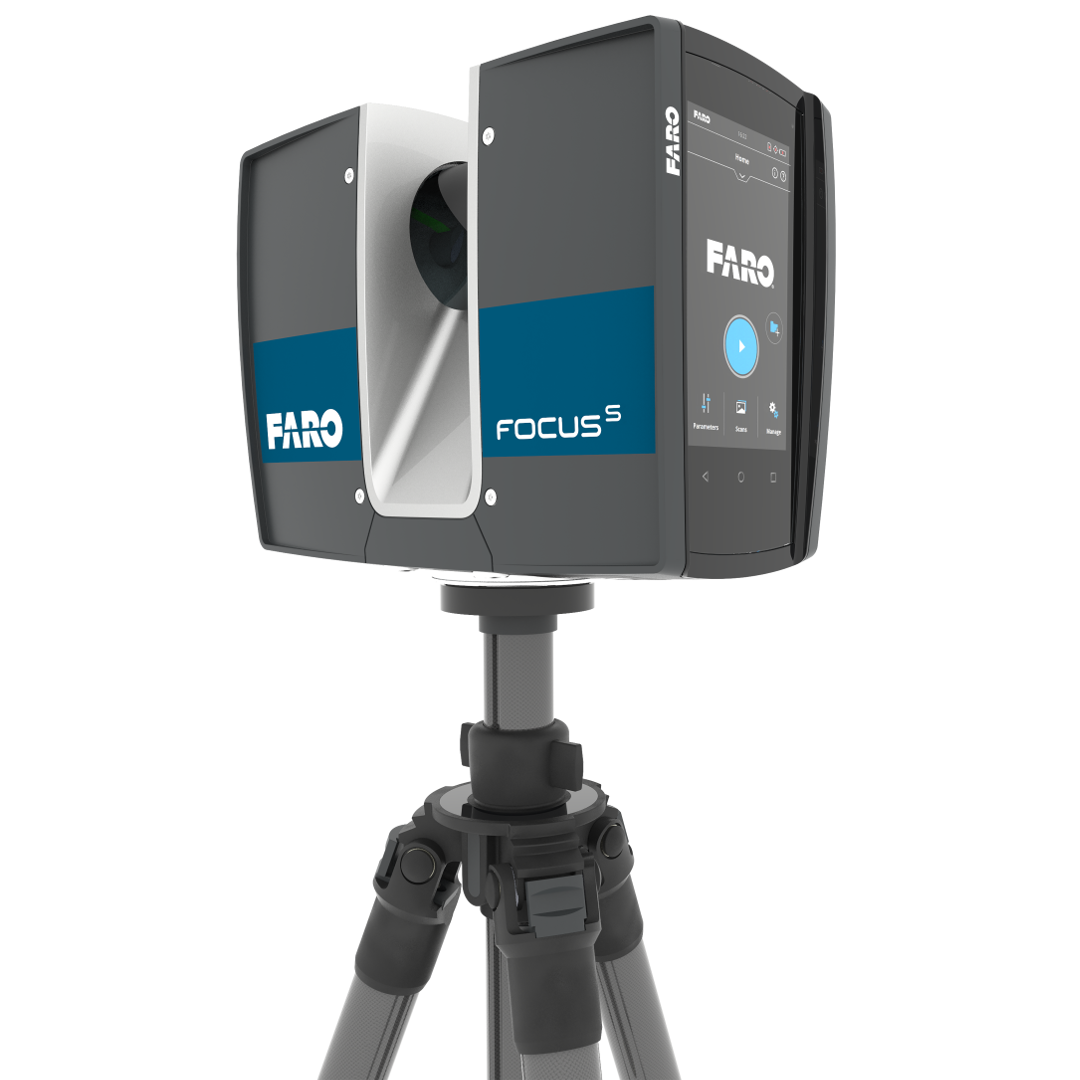}
        \caption{}
        \label{fig:faro_scanner}
    \end{subfigure}
    \hfill
    \begin{subfigure}[b]{0.6\textwidth}
        \centering
        \includegraphics[width=\textwidth]{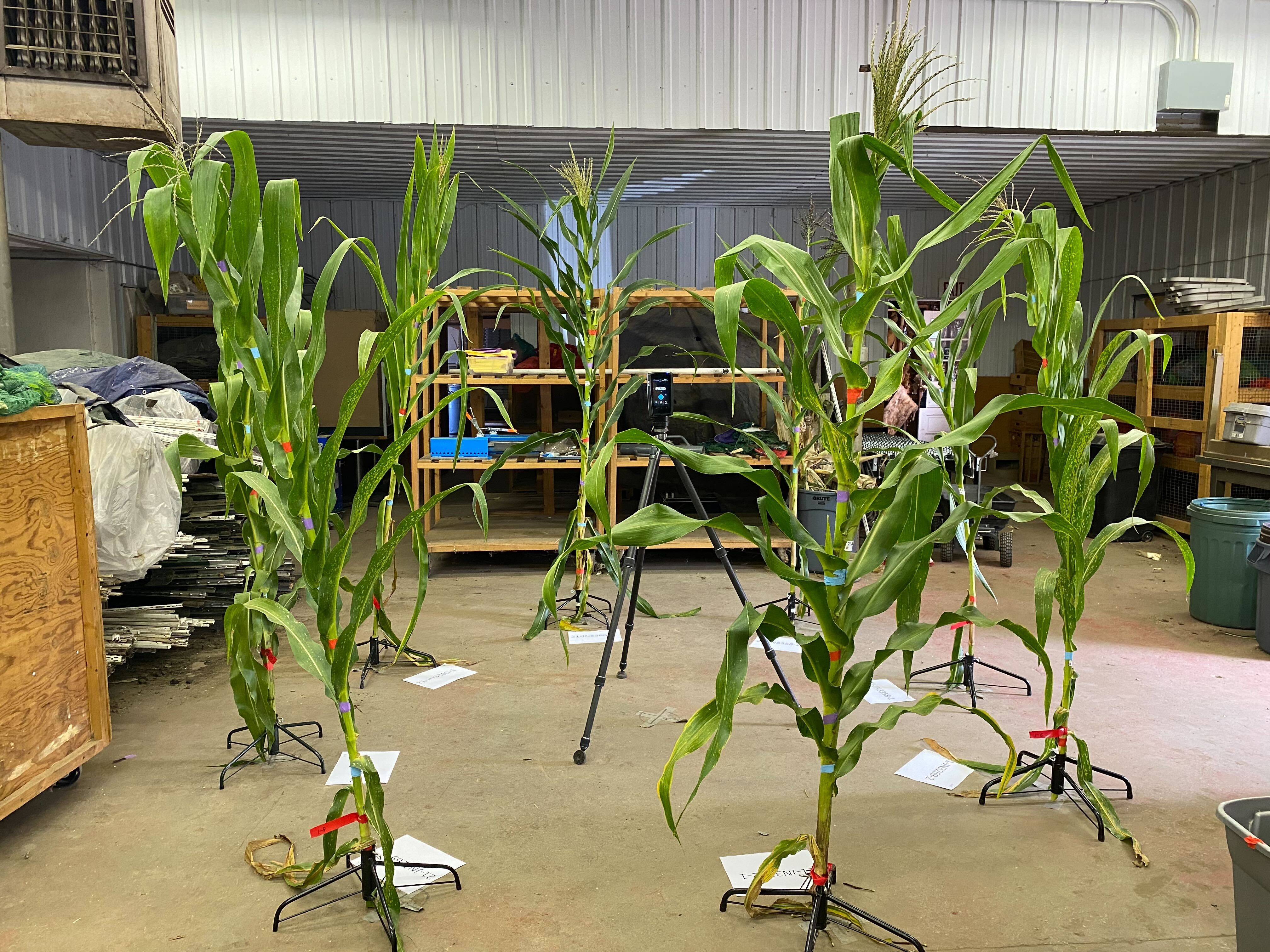}
        \caption{}
        \label{fig:scanning_process}
    \end{subfigure}
    \begin{subfigure}[t]{0.6\textwidth}
        \centering
        \includegraphics[trim=4in 0in 3.9in 0in, clip, width=\linewidth]{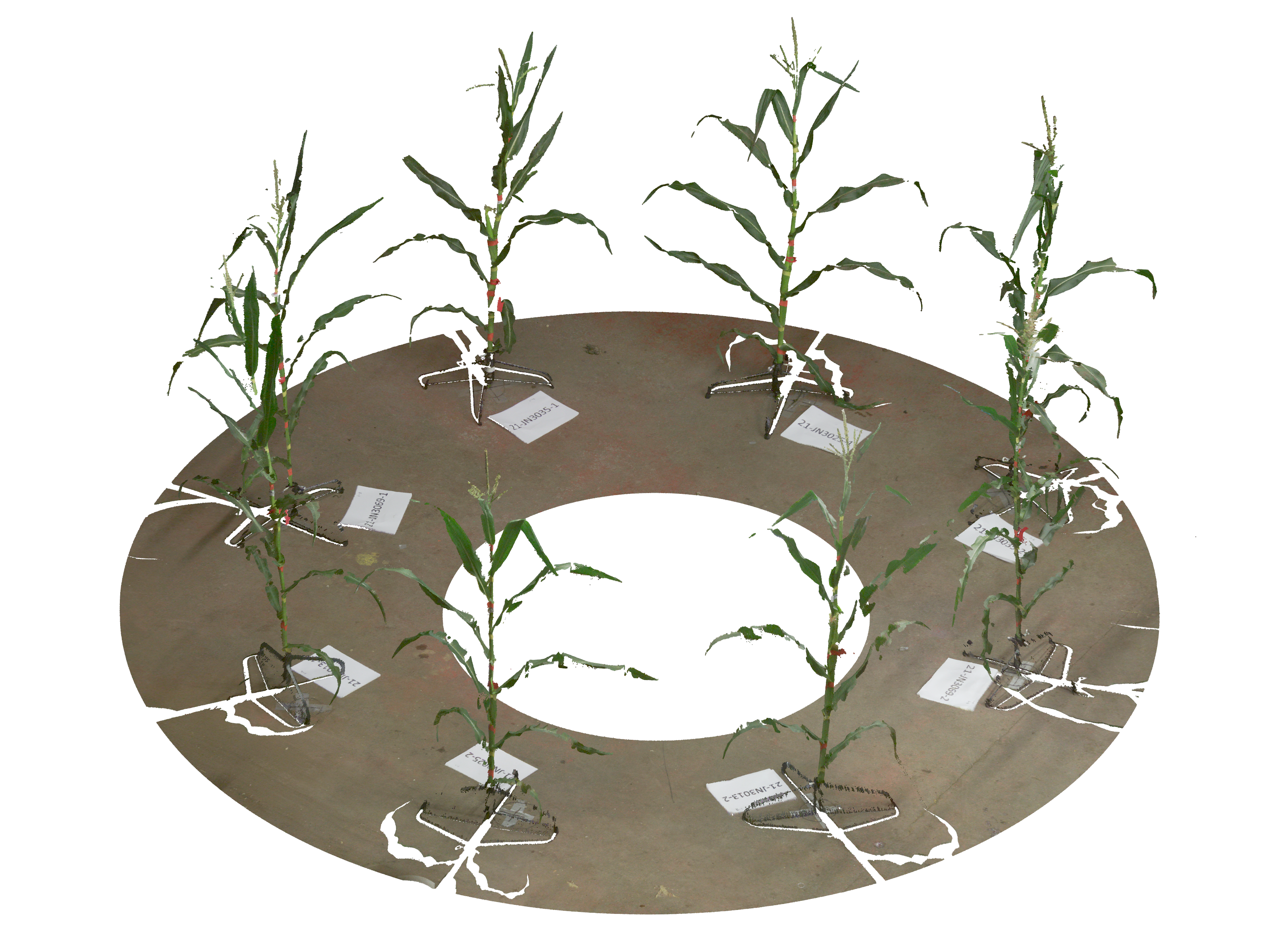}
        \caption{}
        \label{fig:maize_pointcloud}
    \end{subfigure}%
    \hfill
    \begin{subfigure}[t]{0.35\textwidth} 
        \centering
        \includegraphics[trim=5in -7in 5in 0in, clip, width=\linewidth]{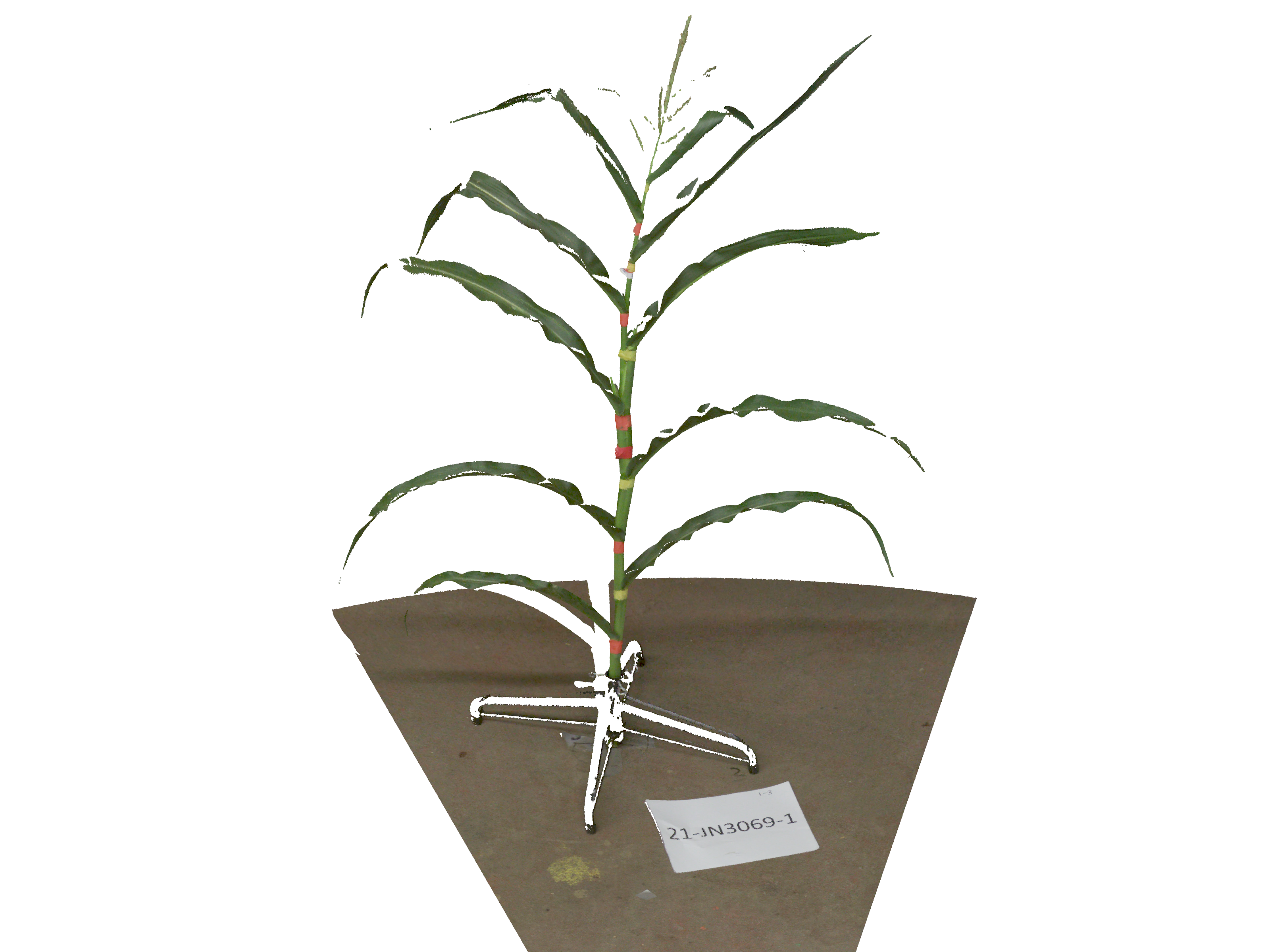}
        \caption{}  
    \end{subfigure}
    \caption{Scanning hardware, setup, and resulting 3D point clouds from the MaizeField3D dataset. (a) The Faro Focus S350 Scanner is used for collecting 3D data of maize plants. This advanced scanner ensures precise and high-resolution point cloud generation, which is critical for phenotypic analysis. (b) Visualization of the scanning process using the Faro scanner. Maize plants are arranged in a circular configuration, allowing simultaneous high-resolution capture of multiple samples in a controlled indoor environment. Point cloud visualization of maize plants are shown in the bottom. (c) The circular arrangement of eight maize plants mounted on a specialized platform, each labeled with unique identifiers. (d) Detailed view of a single maize plant showing characteristic leaf architecture and stem structure.}
    \label{fig:scanner_and_setup}
\end{figure}

\subsection{Data Collection}
 
Field-grown maize plants harvested at about the time of anthesis were selected for data collection from the maize SAM (shoot apical meristem) diversity panel~\citep{leiboff2015genetic} at Iowa State University's Curtis Farm. These lines were planted in 4-row plots at the Iowa State University Woodruff Farm in Ames, Iowa, in the year 2021. Each plot measured 3.04 meters in row length, with plants spaced approximately 15.24 cm apart within rows and 76.2 cm apart between rows, resulting in a planting density of around 84,000 plants per hectare. All plants were collected from the field at the time of anthesis and immediately brought to an indoor environment for imaging and 3D scanning. Each plant was positioned upright using specialized holders to maintain a natural posture, with barcodes scanned to link each plant to its variety tag. Corresponding tag names were printed on white sheets and placed adjacent to the plants for precise labeling and consistent organization.

The 3D data were captured using a Faro Focus S350 Scanner, depicted in \figref{fig:faro_scanner}, which is equipped with an angular resolution of 0.011 degrees, enabling precise point spacing of approximately 1.5 mm at a 10-meter range. The scanner has the capacity to acquire up to 700 million points at a rate of 1 million points per second, ensuring high-resolution outputs suitable for detailed phenotypic analysis. Plants were arranged in batches of eight around the scanner in a circular configuration, as shown in \figref{fig:maize_pointcloud}, allowing simultaneous capture of multiple samples. This setup minimized overlaps between plants and ensured most surfaces were within the scanner’s line of sight. While the scanner setup captured most of each plant’s surface area, the backside of some plants was not directly visible due to the single-scan circular arrangement. This resulted in small gaps in the resulting point clouds. However, the configuration was optimized to capture as much structural detail as possible from a single vantage point, balancing coverage, resolution, and throughput. The scanner height and settings were dynamically adjusted based on plant height and morphology, ensuring that critical structural features, including upper leaves, stalks, tassels, and ears, were precisely captured.
\figref{fig:scanning_process} provides a visual representation of the scanning setup, showcasing the Faro scanner in operation while capturing maize plants arranged in a circular configuration. The controlled indoor environment eliminated external variables such as wind and inconsistent lighting, ensuring uniformity and accuracy across the dataset. Scans were conducted batch by batch until the full dataset was acquired.

\pagebreak

\subsection{Data Processing}

The raw 3D scan data were processed into detailed 3D point clouds using FARO SCENE software, which integrated RGB color values to the outputs. The area of interest, containing only the plants, was then cropped from the scans to exclude extraneous elements. To reduce noise, statistical outlier removal was applied using both global and local point-to-point distance distributions, effectively eliminating artifacts caused by environmental interference or scanner irregularities. This preprocessing ensured high-quality point clouds with precise structural and color details, suitable for downstream phenotypic analyses. We detail these steps further below.

To separate individual plants, CloudCompare software was used. This process, referred to as plant isolation, involved using the crop tool in CloudCompare to extract each plant from the circular arrangement. Identification and naming of the individual plants were facilitated by the paper labels placed in front of each plant during scanning. These labels ensured accurate association between physical plants and their digital representations. Once isolated, each plant was saved as an individual `.ply` file with proper naming conventions. This plant isolation process preserved the structural integrity of the plants while ensuring readiness for downstream analysis.

Following this, a quality control process was implemented. Extraneous elements such as walls, tables, and personnel inadvertently captured during scanning were removed. Each set of maize point clouds was manually inspected to verify its quality, ensuring that plant structures were intact, free from artifacts, and properly cropped. Multiple rounds of review were conducted to maintain accuracy and reliability across the dataset. Additionally, a metadata file was developed to document quality assessments and plant-specific characteristics, including the presence of tassels, maize ears, and resolution. These metadata files provide a useful context for researchers, facilitating targeted analyses and ensuring the dataset’s suitability for diverse applications, including machine learning model training, phenotypic studies, and agricultural research.

\section{Data Annotation}

\subsection{Data Segmentation}

The segmentation of maize plant point cloud data is a key contribution for enabling phenotypic analysis and ML applications. Automated graph-based segmentation methods were combined with manual checks in this procedure. We report precise separation of plant components, including leaves and stalks, while maintaining a high standard of quality.

\subsubsection{Segmentation Workflow}

The initial segmentation was performed using a graph-based approach. This automated process included the following steps:

First, the structural skeleton of the point cloud was extracted through a process known as skeletonization. This skeletonization was performed using the PC Skeletor library~\citep{meyer2023cherrypicker}, which applies Laplacian-based contraction to produce a graph-based representation of the input point cloud. The parameters were adjusted to generalize across different maize examples, effectively identifying the primary stalk and branching patterns. Following this, stalk node identification was performed to detect the top and bottom nodes of the maize stalk based on z-coordinate extremes, with a center bias parameter applied to prioritize median x and y coordinates.

Next, the shortest path in the maize skeleton between the identified top and bottom nodes was calculated. Nodes along this path were classified as stalk nodes, while other nodes were classified as leaf nodes. To enhance segmentation accuracy, stalk nodes were temporarily removed from the skeleton graph.

The segmentation of leaves was then performed by identifying each connected component remaining in the skeleton graph after the removal of stalk nodes. Connected components with a small number of nodes were excluded to avoid noise. Subsequently, each point in the maize point cloud was assigned to its corresponding leaf or stalk based on proximity to these connected components.

Finally, points were colored with RGB values corresponding to their associated leaf or stalk, generating a point cloud file that visually represents the segmented maize plant.

\subsubsection{Manual Inspection and Quality Control}

Following the graph based process, we performed manual inspection and editing of the post-processed data. This manual refinement process was conducted using a Python-based preprocessing script in combination with the \textit{CloudCompare (CC)} application. 

Initially, each leaf and the stalk were extracted into separate \texttt{.ply} files for individual inspection and correction. This separation allowed for focused evaluation and adjustment of specific plant components. During this phase, common segmentation errors were addressed. For instance, misclassified points, such as parts of a leaf erroneously grouped with the stalk, were carefully segmented out and merged with the correct component. Similarly, interleaf overlaps, where points from one leaf were mistakenly grouped with a neighboring leaf, were rectified by accurately reassigning those points.
To maintain consistency, the leaf numbering was reviewed and reordered from the bottom to the top of the plant. This step ensured a clear structure in the dataset.

\subsubsection{Recoloring and Merging}

Following segmentation, a recoloring and merging process was implemented. Each plant component was recolored to adhere to a consistent color scheme. For example, the first leaf in every plant was assigned the same color, followed by the second leaf, and so on, with the stalk uniformly colored across all samples. This uniformity enhanced the dataset's usability and interoperability. The color scheme includes labels for up to 16 leaves, which reflects the maximum number of segmented leaves in our segmented dataset. So we limit the color encoding to 16 leaf labels and the stalk. \tabref{tab:color_coding} in the Supplementary Materials details the color ordering of the leaves and stalks.

After recoloring, the corrected components were merged into a single \texttt{.ply} file for each plant. This consolidated format preserved the color labels, allowing for easy identification of individual leaves and the stalk within each plant. The merged files provided a cohesive representation of the plant structure while retaining the detailed segmentation information.

\subsubsection{Segmented Dataset}

The segmentation and refinement processes were applied to \textbf{520 plants}. \figref{fig:segmented_plants} showcases three examples of the segmented maize plant point clouds, highlighting the color-coded segmentation data among the 520 segmented plants in the dataset. Additional examples are provided in \figref{fig:additional_segmented_plants}.

\subsubsection{Procedural Model}
\label{procedural}
\begin{figure}[t!]
    \centering

    \begin{subfigure}[b]{0.21\linewidth}
        \centering
        \includegraphics[trim=1cm 2cm 1cm 2cm, clip, width=\linewidth]{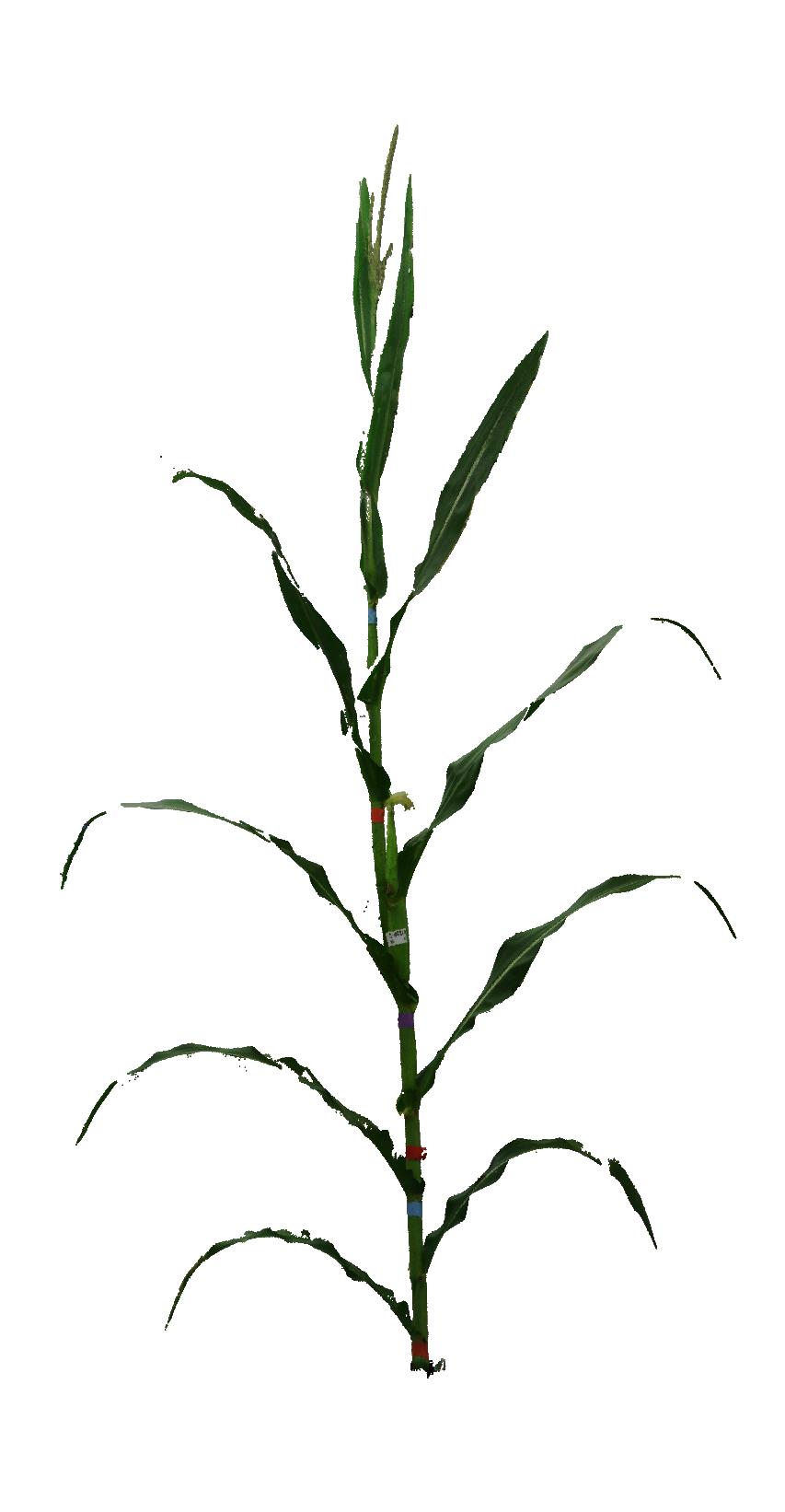}
        \caption{}
    \end{subfigure}
    \hfill
    \begin{subfigure}[b]{0.21\linewidth}
        \centering
        \includegraphics[trim=1cm 2cm 1cm 2cm, clip, width=\linewidth]{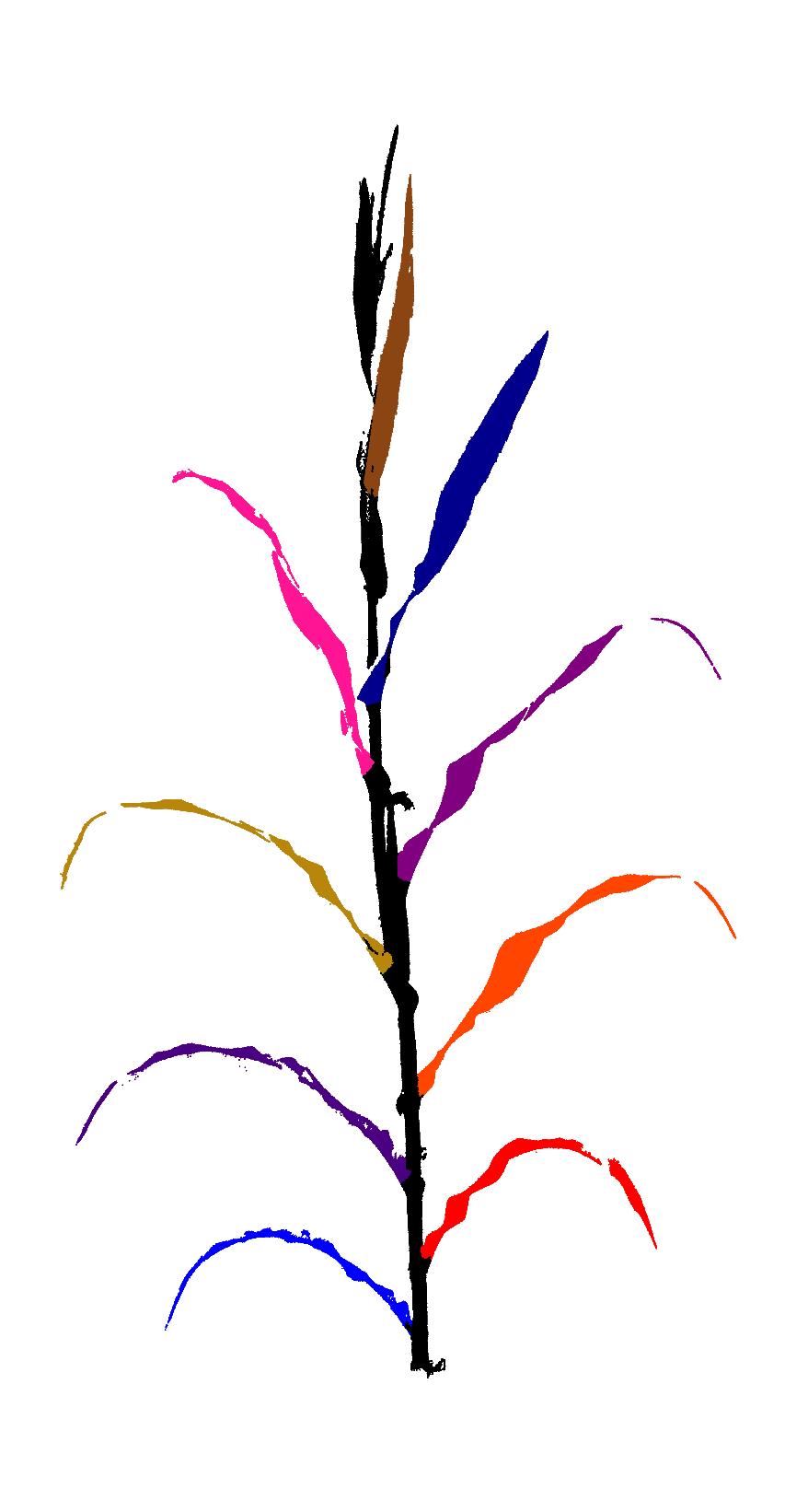}
        \caption{}
    \end{subfigure}
    \hfill
    \begin{subfigure}[b]{0.21\linewidth}
        \centering
        \includegraphics[trim=1cm 2cm 1cm 2cm, clip, width=\linewidth]{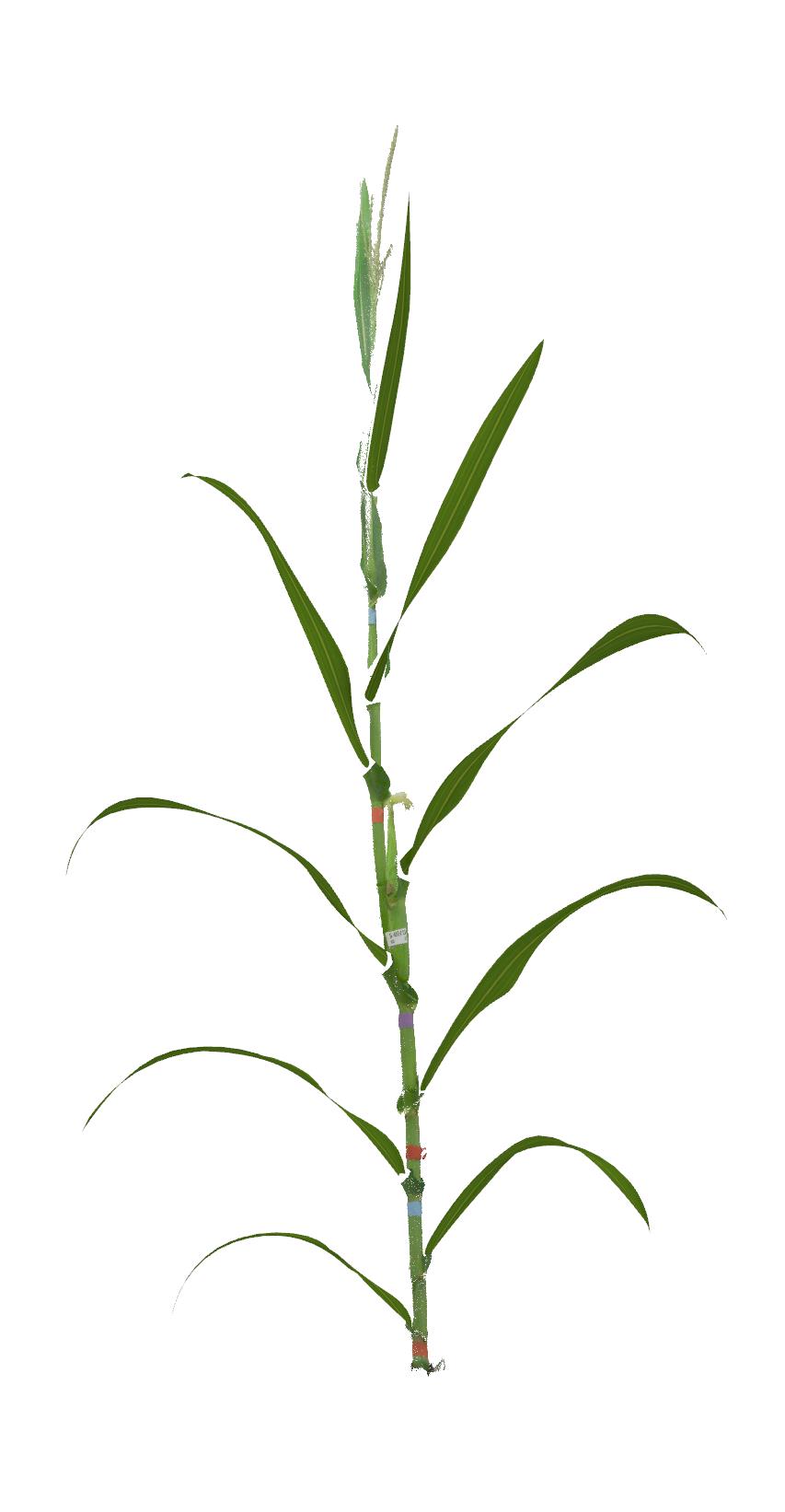}
        \caption{}
    \end{subfigure}

    \vspace{1em}

    \begin{subfigure}[b]{0.21\linewidth}
        \centering
        \includegraphics[trim=1cm 4cm 1cm 2cm, clip, width=\linewidth]{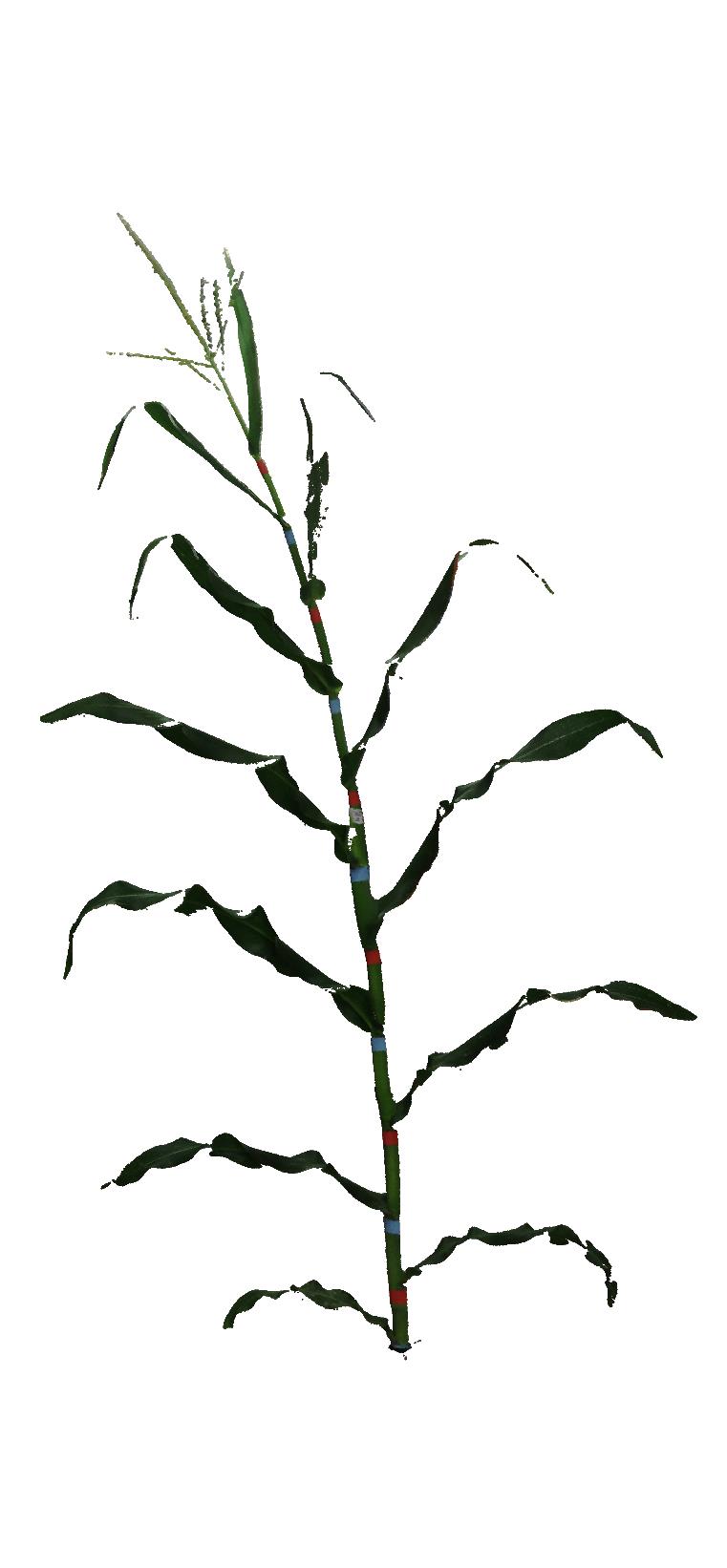}
        \caption{}
    \end{subfigure}
    \hfill
    \begin{subfigure}[b]{0.21\linewidth}
        \centering
        \includegraphics[trim=1cm 4cm 1cm 2cm, clip, width=\linewidth]{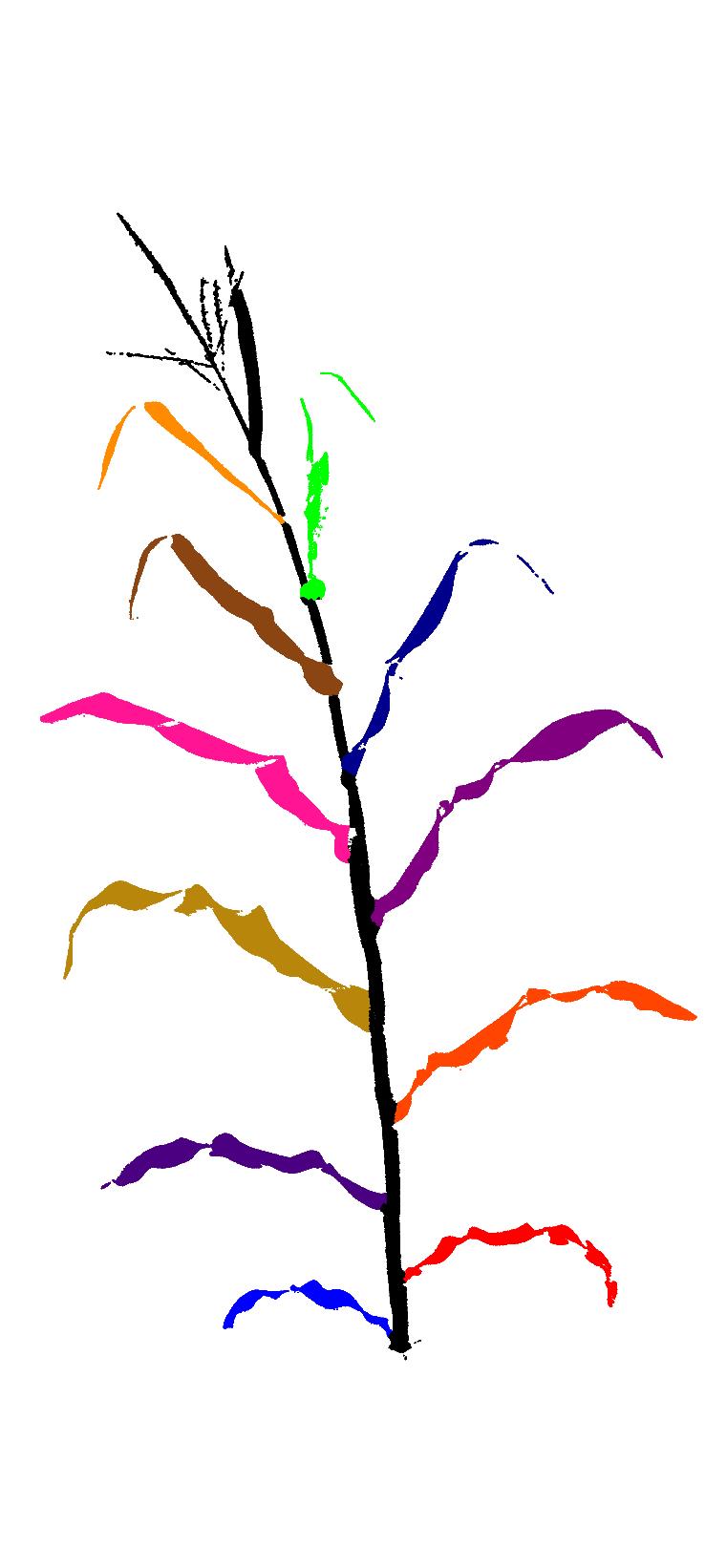}
        \caption{}
    \end{subfigure}
    \hfill
    \begin{subfigure}[b]{0.21\linewidth}
        \centering
        \includegraphics[trim=1cm 4cm 1cm 2cm, clip, width=\linewidth]{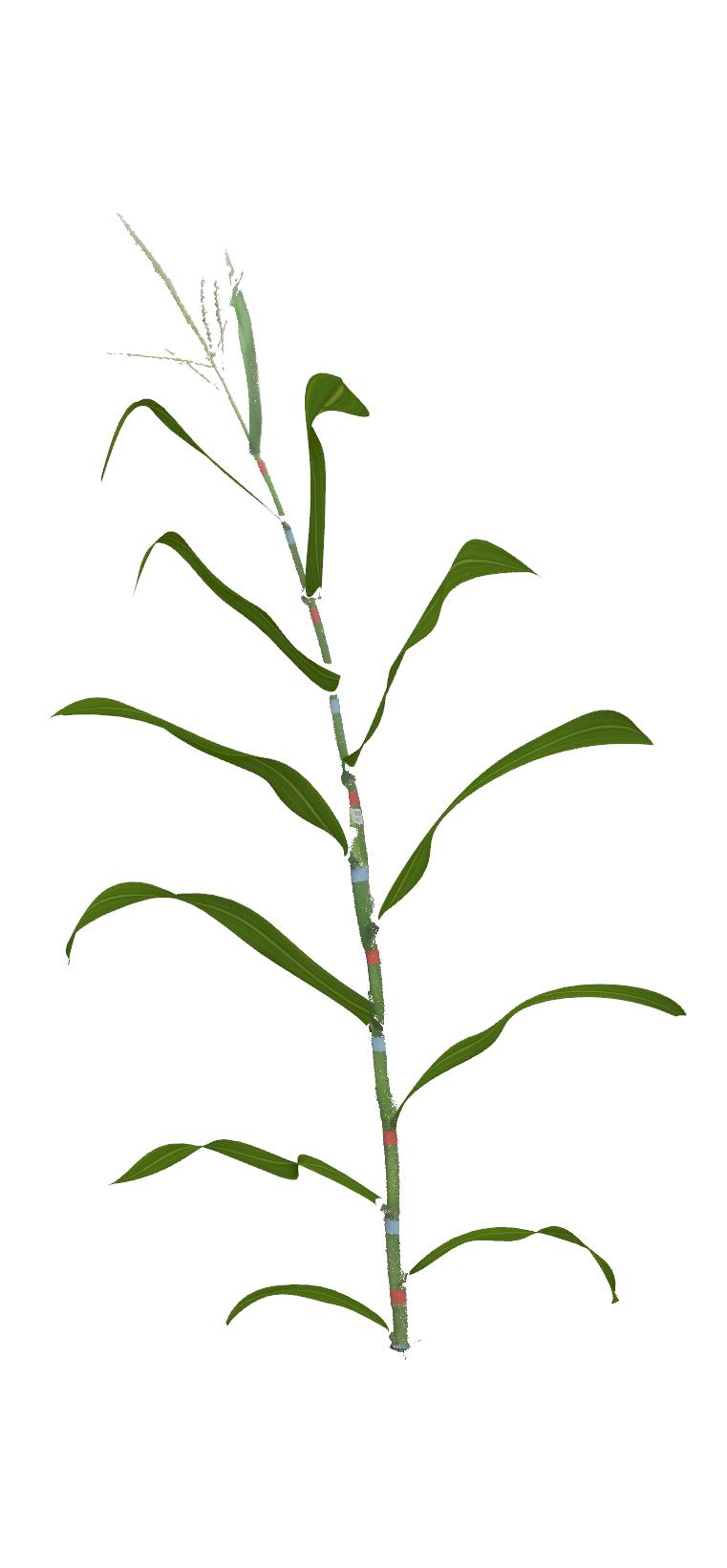}
        \caption{}
    \end{subfigure}

    \caption{Processing of two maize plants: B73 (top row) and Mo17 (bottom row). (a, d) Raw data, (b, e) segmented data, and (c, f) final NURBS surfaces.}
    \label{fig:B73_Mo17}
\end{figure}

Procedural modeling is a computational approach to generating structured 3D representations from raw data by applying algorithmic rules and mathematical functions. Unlike raw point clouds, which are unstructured and can be noisy or incomplete, procedural models create parametric surfaces that provide a compact, scalable, and editable representation of plant structures. This method enables precise reconstructions that preserve geometric accuracy while allowing for controlled variations, making it particularly useful for synthetic data augmentation, structural analysis, and phenotyping applications. Procedural modeling is especially valuable for plant phenotyping, where capturing complex leaf curvature and growth patterns is essential for trait analysis and computational simulations.
The segmented dataset of 520 maize plant point clouds was used as input for a procedural modeling approach developed in Hadadi et al.~\citep{hadadi2025procedural}. This method enabled the high-fidelity reconstruction of 3D surfaces that accurately capture the shape and structure of individual maize leaves. By leveraging this procedural model, each segmented plant was processed to NURBS-based surfaces. The procedural model ensured that the reconstructed surfaces were closely aligned with the original segmented point clouds. \figref{fig:B73_Mo17} illustrate examples of these reconstructions, showcasing the transition from raw point cloud data to the final NURBS surface representation.

To enhance the usability of this dataset, the reconstructed outputs are made publicly available in the following formats:
\begin{itemize}
    \item \textbf{STL Files:} Containing 3D reconstructed surfaces, suitable for visualization and simulation applications.
    \item \textbf{DAT Files:} Providing control point data for the NURBS surfaces, allowing further refinements and integration into CAD workflows.
\end{itemize}

\begin{figure}[t!]
    \centering
    \begin{subfigure}[b]{0.19\linewidth}
        \centering
        \includegraphics[trim=1cm 6cm 1cm 2.2cm, clip, width=\linewidth]{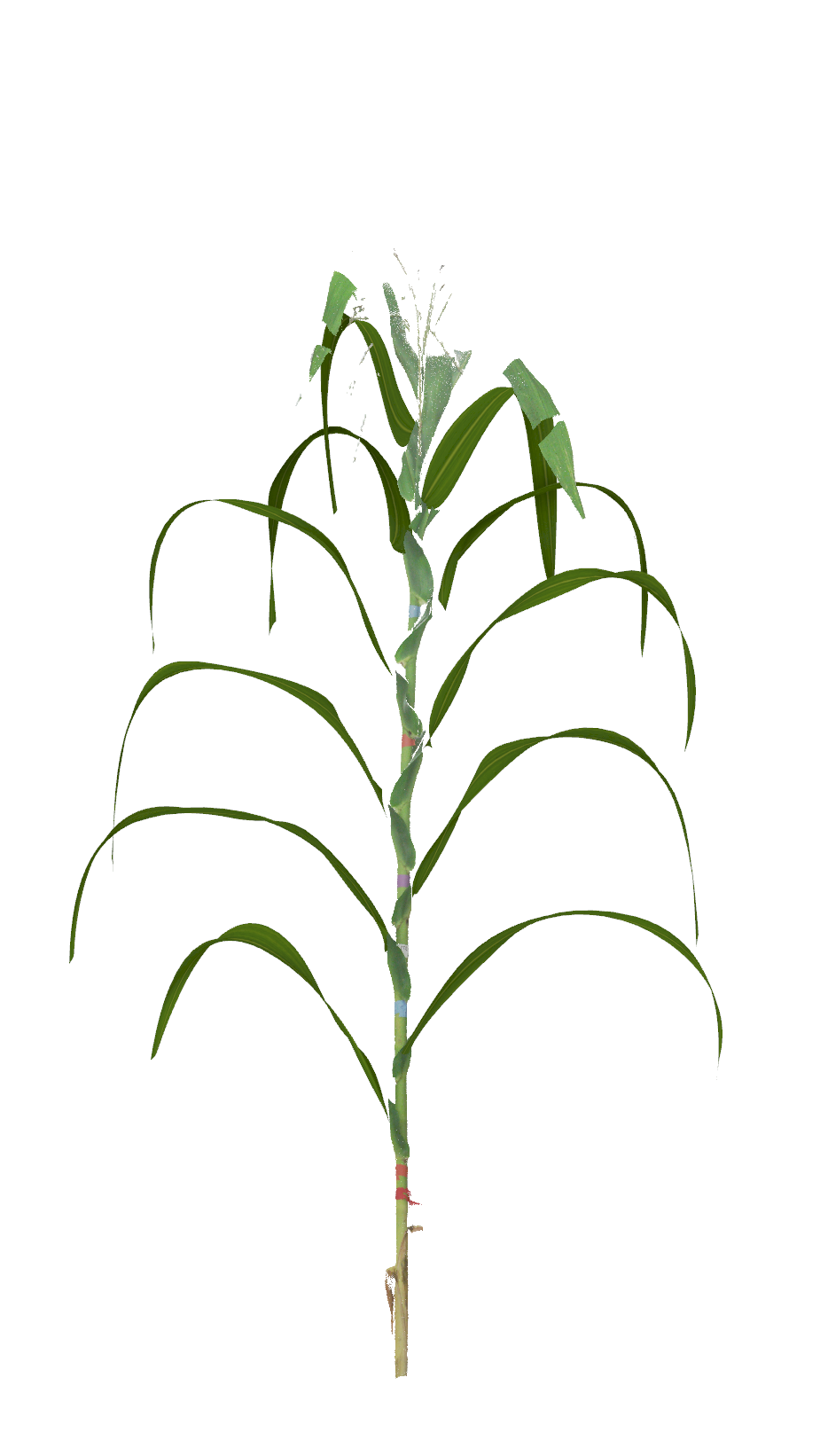}
        \caption{}
    \end{subfigure}
    \hfill
    \begin{subfigure}[b]{0.19\linewidth}
        \centering
        \includegraphics[trim=2cm 6.5cm 1cm 2.0cm, clip, width=\linewidth]{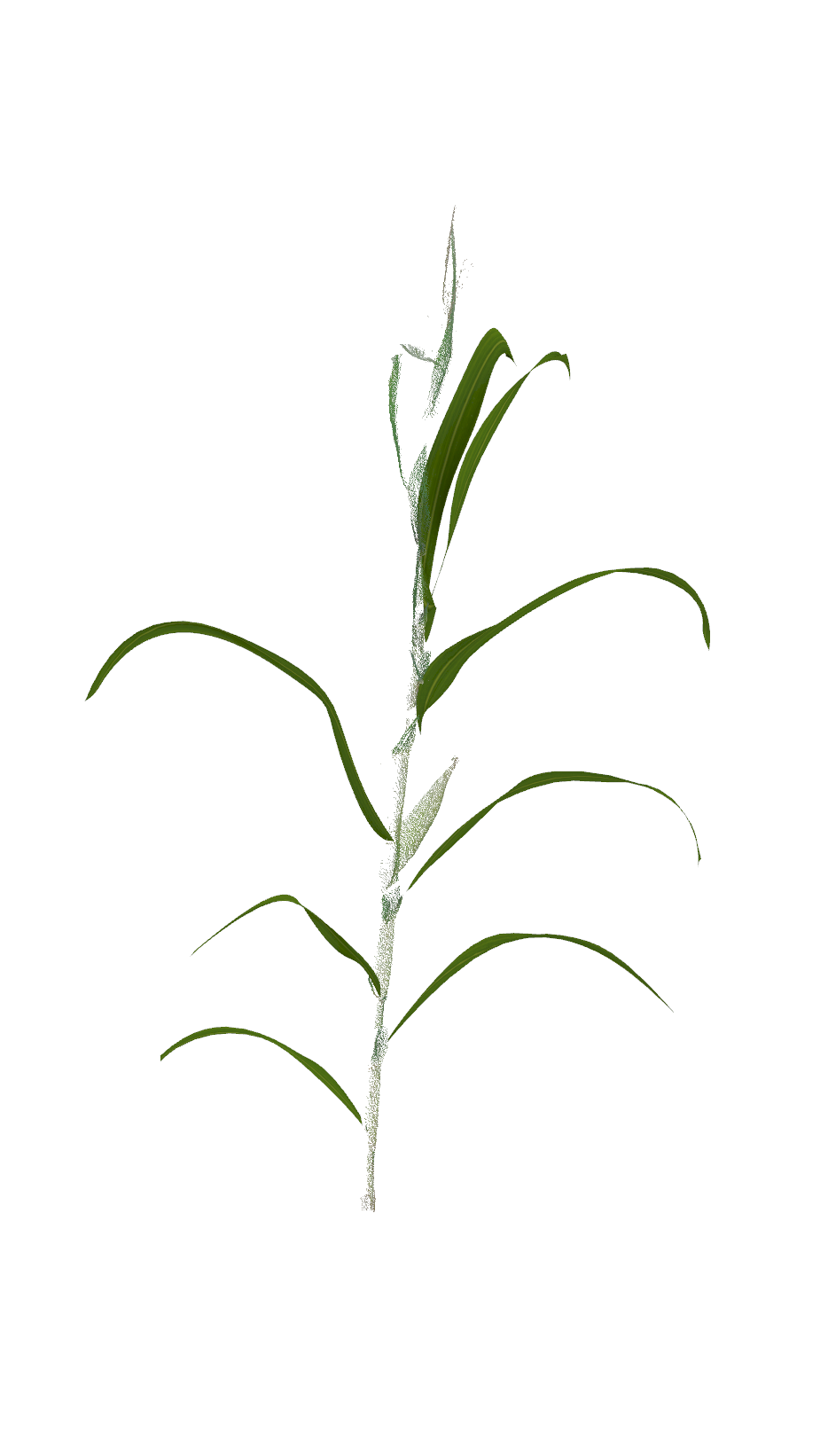}
        \caption{}
    \end{subfigure}
    \hfill
    \begin{subfigure}[b]{0.19\linewidth}
        \centering
        \includegraphics[trim=1cm 4cm 1cm 6.1cm, clip, width=\linewidth]{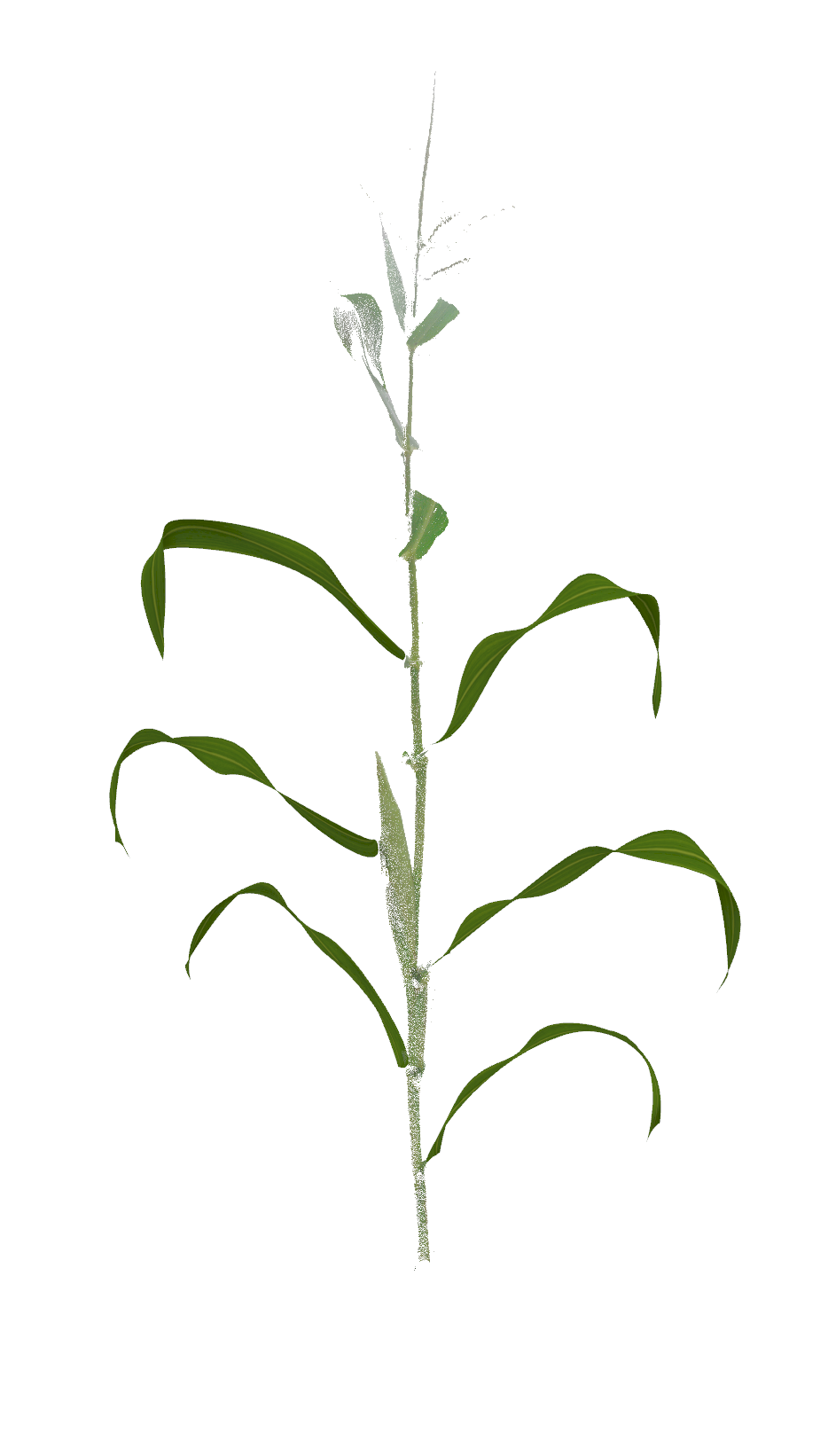}
        \caption{}
    \end{subfigure}
    \hfill
    \begin{subfigure}[b]{0.19\linewidth}
        \centering
        \includegraphics[trim=1cm 2cm 1cm 6.1cm, clip, width=\linewidth]{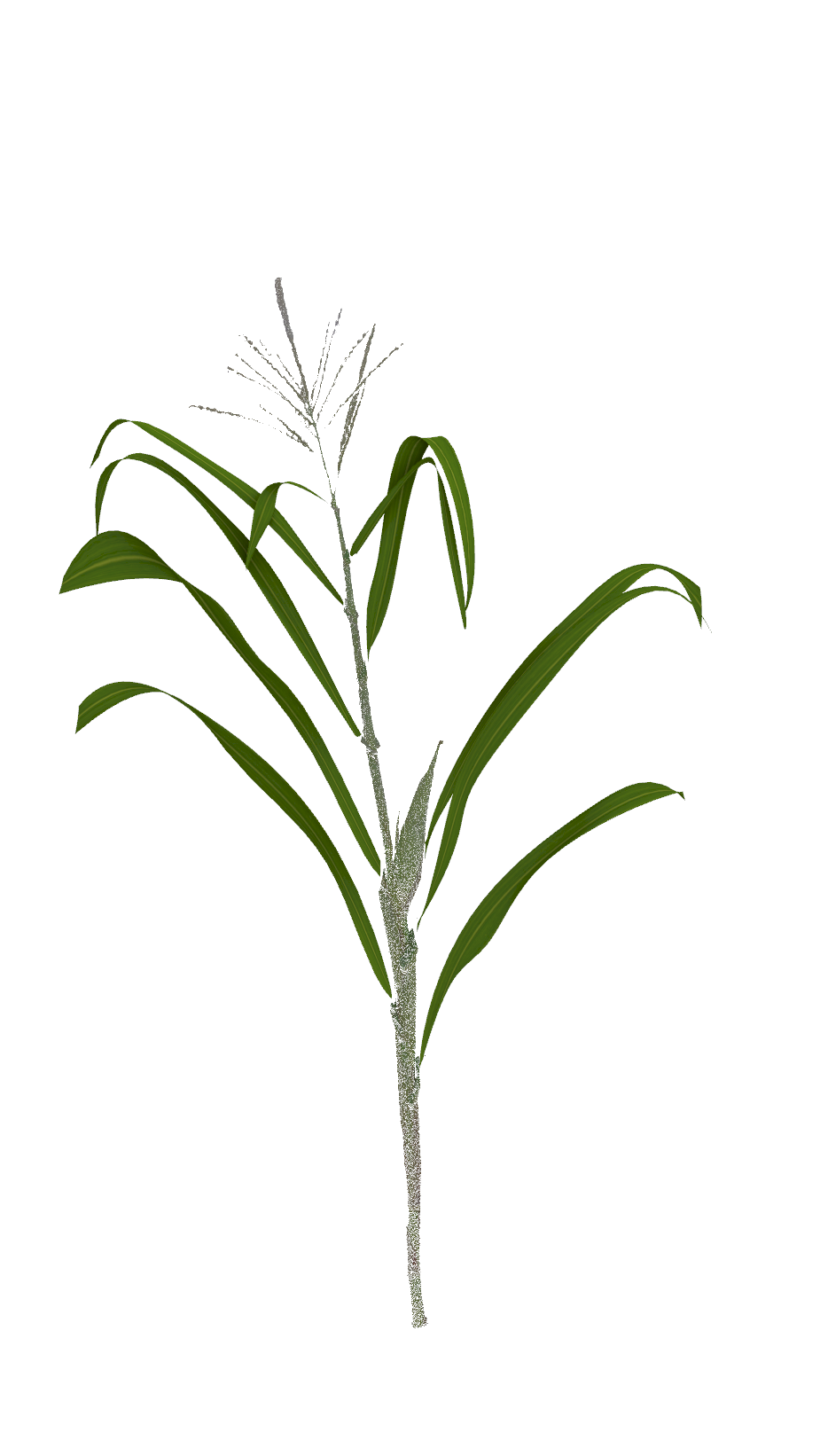}
        \caption{}
    \end{subfigure}
    \hfill
    \begin{subfigure}[b]{0.19\linewidth}
        \centering
        \includegraphics[trim=7cm 0cm 7cm 0cm, clip, width=\linewidth]{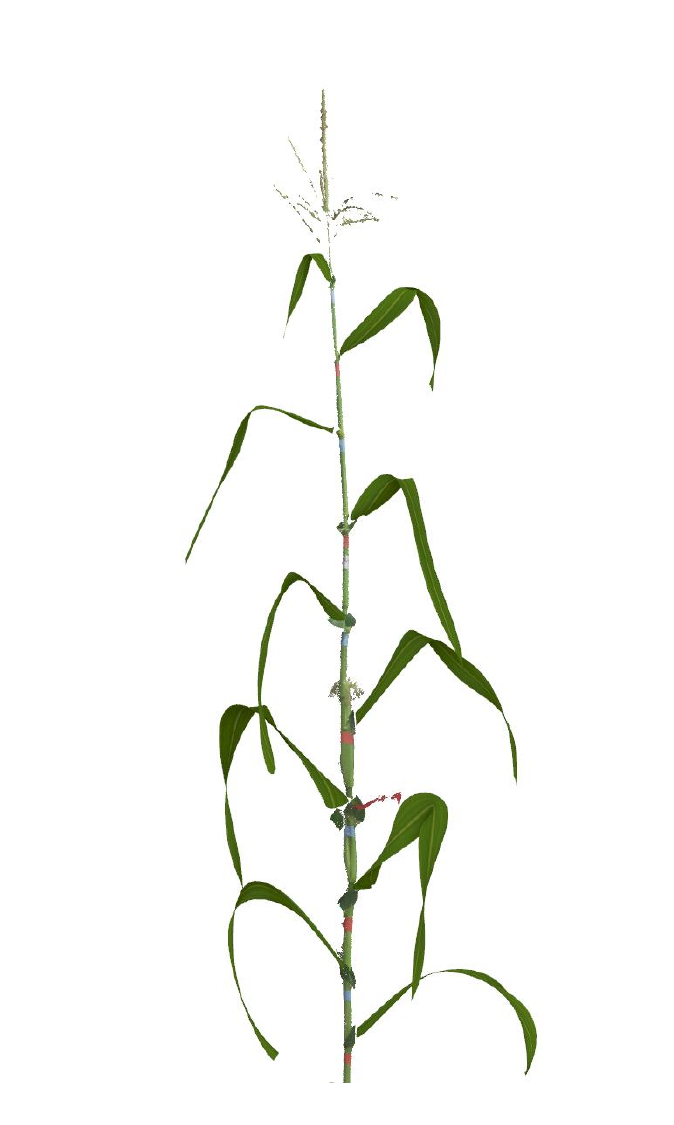}
        \caption{}
        \label{fig:plant_e}
    \end{subfigure}
    \caption{NURBS surface reconstruction using the procedural model for five diverse maize plants.}
    \label{fig:procedural_output}
\end{figure}

These additional files complement the segmented point clouds, bridging raw data with high-fidelity surface models that can be utilized for diverse research applications. \figref{fig:procedural_output} illustrates the NURBS surface reconstructions derived from the procedural modeling approach applied to four distinct maize plants. The examples highlight the diversity in maize leaf structures across maize plant varieties. The process accurately captures the complex geometries of maize leaves, including their curvature and edges.

\section{Data Downsampling}
Upon analyzing the dataset, substantial variability was observed in the number of points per point cloud. The point counts ranged from as low as 150K points to as high as 2.6 million points, with most point clouds centered around 600K points. This variability, while reflective of the diversity in plant structures, may pose challenges for downstream tasks requiring uniform point counts.

\begin{figure}[t!]
    \centering
    \includegraphics[width=0.95\textwidth]{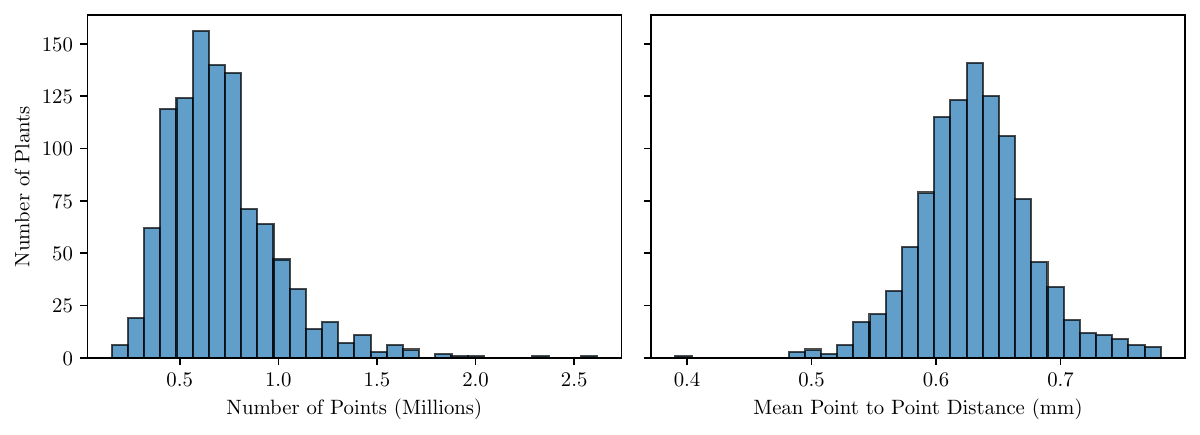}
    \caption{Histograms summarizing the original point cloud dataset before downsampling. The left histogram illustrates the distribution of the number of points per plant, scaled to millions, highlighting the variability in point cloud sizes across the dataset. The right histogram shows the distribution of mean nearest neighbor distances, converted to millimeters, capturing the spatial density of points within each plant's point cloud. Together, these histograms provide insights into the scale and resolution of the dataset, emphasizing its diversity and structural fidelity.}
    \label{fig:histograms_orginial}
\end{figure}

~\figref{fig:histograms_orginial} provides an overview of the original point cloud dataset before any downsampling or modifications. The histograms highlight two key aspects of the dataset:

\begin{itemize}
    \item \textbf{Point Count Distribution (Left Plot):} The first histogram depicts the distribution of the number of points per plant, scaled to millions. The dataset exhibits a wide range of point counts across the plants, reflecting the diversity in plant structure and size. Most plants have a point count within a specific range, with fewer plants having extremely high or low counts.

    \item \textbf{Point to Point Distance (Right Plot):} The second histogram shows the distribution of the mean nearest neighbor distances, converted to millimeters. This measure captures the spatial density of points within each plant's point cloud. A narrower spread in this distribution indicates relatively consistent point density across the dataset.
\end{itemize}

The left panel demonstrates the scale and resolution of the original dataset, while the right panel highlights the structural fidelity retained in the point cloud representation. Such metrics are crucial for understanding how well the dataset represents real-world plant morphology and for determining appropriate preprocessing steps for downstream applications.

To address this variability, downsampled versions of the dataset were created with fixed point counts \(N\). These standardized datasets ensure compatibility with computational frameworks and machine-learning pipelines that require consistent input sizes. Additionally, downsampling reduces computational overhead, broadening the dataset’s usability for researchers with varying computational resources. Three target point counts were chosen to meet diverse research needs, including \(100,000\), \(50,000\), and \(10,000\) points. These resolutions accommodate applications ranging from resource-intensive tasks requiring high-resolution data to computationally efficient analyses.

\begin{figure}[t!]
    \centering
    \begin{subfigure}[b]{0.95\textwidth}
        \centering
        \includegraphics[width=0.85\textwidth,trim=0 50 0 0, clip]{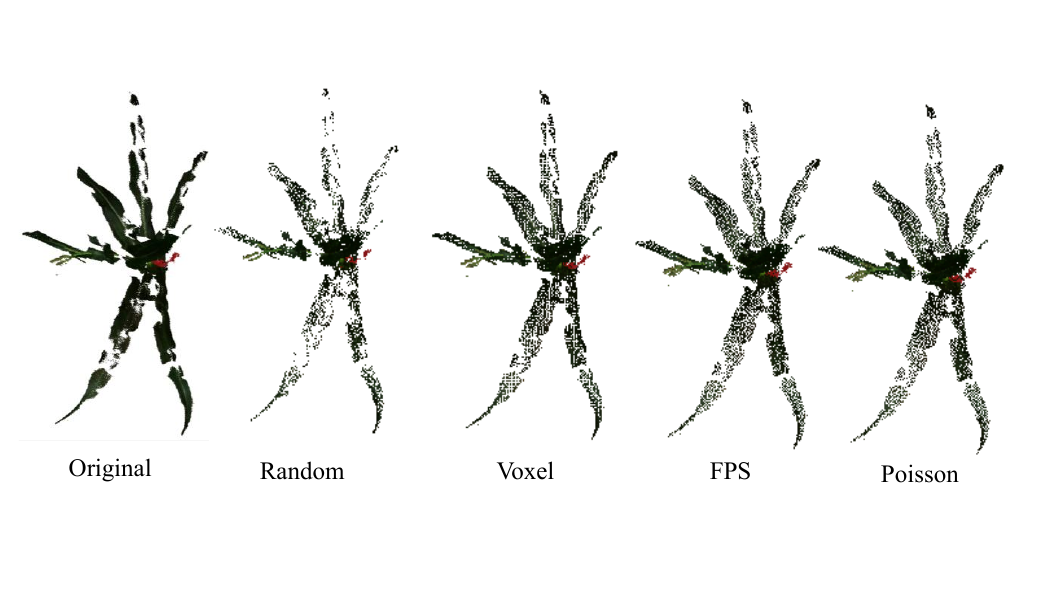}
        \caption{}
        \label{fig:sampling_qualitative}
    \end{subfigure}
    \begin{subfigure}[b]{0.95\textwidth}
        \centering
        \includegraphics[width=0.95\textwidth]{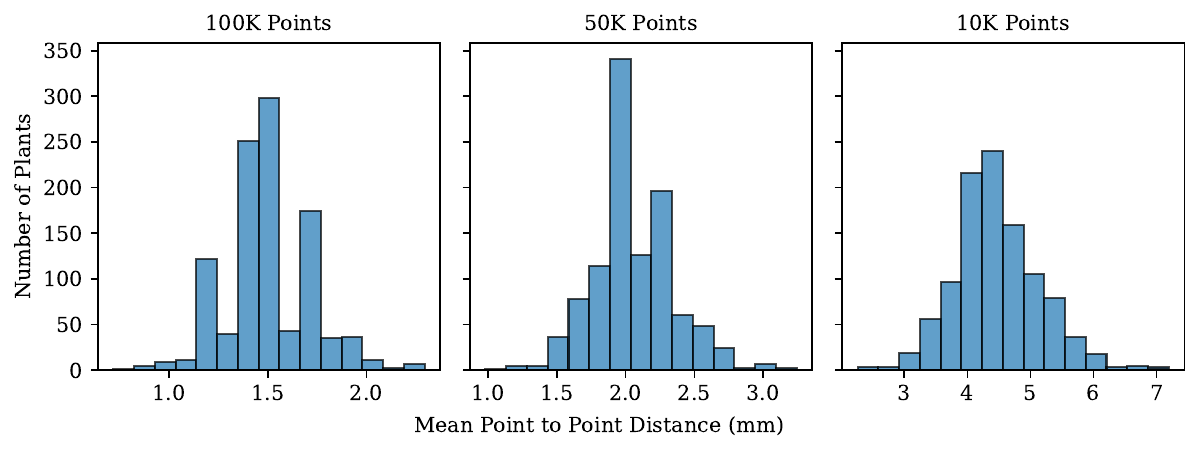}
        \caption{}
        \label{fig:combined_distances}
    \end{subfigure}
    \caption{(a) Qualitative comparison of downsampling strategies on representative point clouds. (b) Comparison of histograms showing the mean point to point distances for the dataset downsampled to 100K points, 50K points and 10K points.}
    \label{fig:sampling_combined}
\end{figure}

Several downsampling techniques were evaluated to identify the most suitable method for downsampling. We used Open3D~\citep{zhou2018open3d} for point cloud processing, including voxel grid downsampling (Voxel), Poisson disk sampling (Poisson), farthest point sampling (FPS), and random downsampling (Random). A comparison was performed for all four methods at three target resolutions—100k, 50k, and 10k points—to evaluate their runtime and output characteristics. As expected, Random downsampling was the fastest, although it lacked spatial consistency. FPS and Poisson Disk Sampling provided more uniform point distributions but incurred significantly higher computational costs. Voxel Grid sampling offered a balance, producing structured outputs with moderate runtime. \figref{fig:sampling_qualitative} qualitatively compares point distributions, supporting the effectiveness of our approach. \tabref{tab:downsampling_comparison} in Supplementary Materials summarizes the runtime performance of each method and

To balance efficiency and structural regularity, we adopted a two-step approach: Voxel Grid downsampling is first applied to preserve plant structure while reducing point count. The voxel size is iteratively increased until slightly over the target size \(N\), followed by Random sampling to reach exactly \(N\) points. This hybrid method ensures spatial uniformity with low computational cost.

 We have provided a Python script to allow researchers to generate custom downsampled datasets with any desired point count \(N\). \figref{fig:combined_distances} presents the histograms of mean nearest neighbor distances for the three downsampled versions. As the point count decreases, the mean point-to-point distances increase, reflecting the reduced resolution. These visualizations highlight the trade-off between dataset size and structural detail, enabling researchers to choose the resolution best suited to their computational and analytical needs.

\subsection{Metadata}

The metadata file contains the following columns:

\begin{itemize}
    \item \texttt{TagName}: The unique identifier for the plant, associated with its genotype through a separate look-up table.
    \item \texttt{IsConnected}: Indicates if the plant's stem and leaves are connected with no gaps (\textit{Yes/No}).
    \item \texttt{IsEasyToCountLeaves}: Categorizes the ease of counting leaves as \textit{easy}, \textit{moderate}, or \textit{hard}.
    \item \texttt{AreLeavesTouching}: Specifies if the leaves of the plant are touching each other (\textit{Yes/No}).
    \item \texttt{HasColor}: Indicates the presence of RGB color data in the point cloud (\textit{Yes/No}).
    \item \texttt{HasEar}: Indicates whether the plant has an ear (\textit{Yes/No}).
    \item \texttt{HasTassel}: Indicates whether the plant has a tassel (\textit{Yes/No}).
    \item \texttt{Size}: Specifies the number of points in the point cloud.
    \item \texttt{Comment}: Includes any additional notes or observations relevant to the plant or the quality of its point cloud.
\end{itemize}

\begin{figure}[t!]
\centering
\begin{minipage}{0.35\textwidth}
    \centering
    \includegraphics[trim=9in 0in 8in 0in, clip, width=\textwidth]{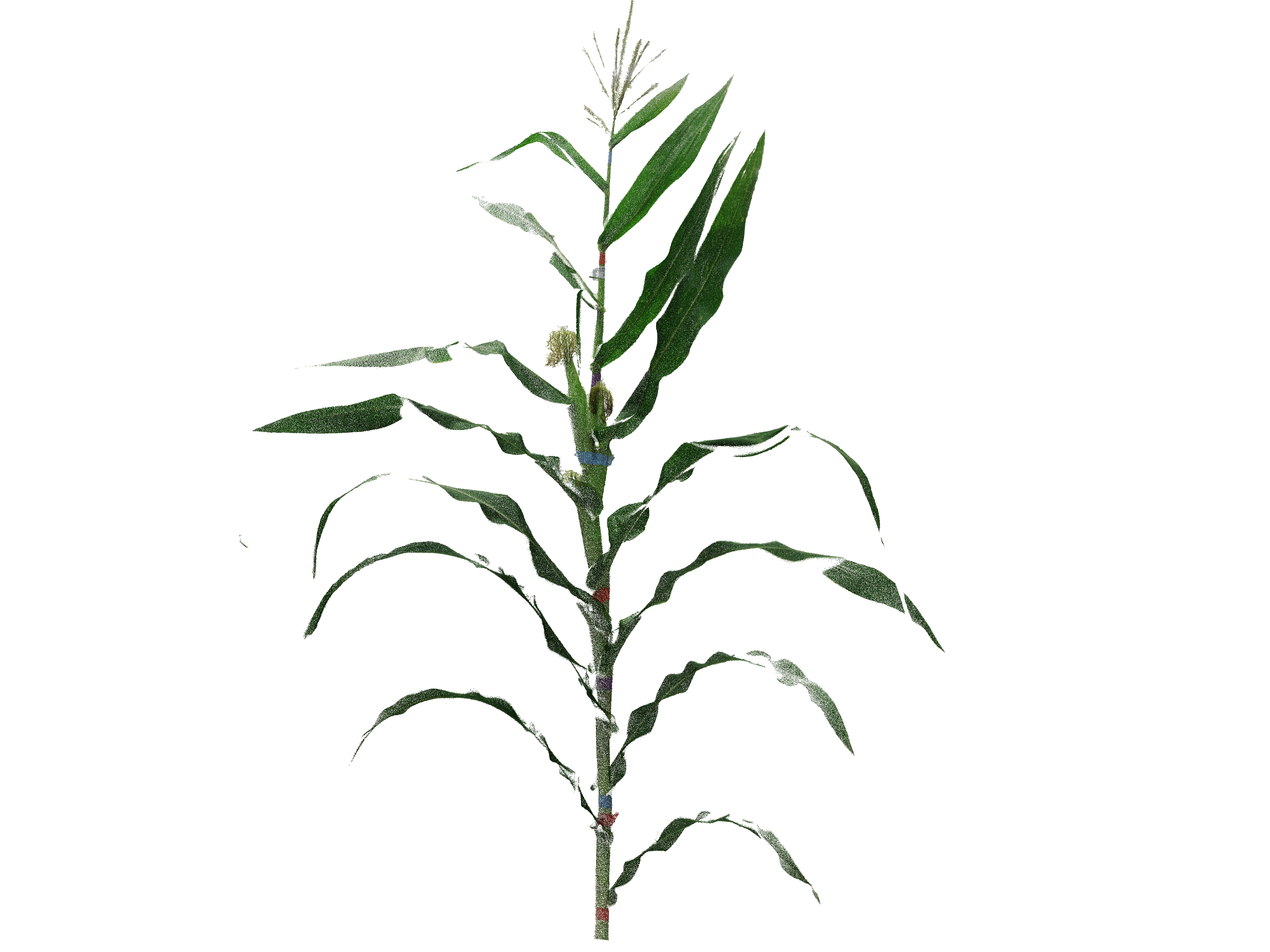}
\end{minipage}%
\begin{minipage}{0.45\textwidth}
    \centering
    \renewcommand{\arraystretch}{1.5}
    \label{tab:pointcloud_metadata}
    \resizebox{\textwidth}{!}{%
    \begin{tabular}{ll}
    \toprule
    \textbf{Column Name} & \textbf{Value} \\
    \midrule
    \texttt{TagName} & 0433 \\
    \texttt{IsConnected} & Yes \\
    \texttt{IsEasyToCountLeaves} & Easy \\
    \texttt{AreLeavesTouching} & No \\
    \texttt{HasColor} & Yes \\
    \texttt{HasEar} & Yes \\
    \texttt{HasTassel} & Yes \\
    \texttt{Size} & 1,558,177 points \\
    \texttt{Comment} & This is a good plant. \\
    \bottomrule
    \end{tabular}%
    }
\end{minipage}
\caption{Side-by-side representation of a maize plant point cloud and its corresponding metadata table. The point cloud is visualized on the left, while the table on the right provides detailed metadata annotations, enabling researchers to interpret the dataset effectively for downstream analyses.}
\label{fig:pointcloud_metadata}
\end{figure}

The metadata provides a detailed description of the structural and phenotypic characteristics of each plant (see \figref{fig:pointcloud_metadata}). In addition, it enables targeted data selection. For example, researchers can quickly identify plants with a specific number of leaves or those with complete color information to suit their experimental requirements.


\section{Potential Applications}

The curated 3D point cloud dataset of maize plants offers numerous opportunities for advancing agricultural research and crop improvement. We identify some promising applications below.

\paragraph{Feature Extraction and Phenotypic Analysis:}
The dataset's detailed segmentation and annotations support the extraction of critical phenotypic traits, including the number of leaves, leaf length, width, and curvature, as well as stalk dimensions. These features can be used to quantify plant architecture, model growth patterns, and analyze genotype-phenotype relationships~\citep{Tardieu2017}. The accurate segmentation of leaves and stalks further facilitates studies on plant competition, light interception, and resource allocation within the canopy~\citep{Araus2018}.

\paragraph{Shape Representation and 3D Modeling:}
The high-resolution 3D point clouds serve as valuable benchmarks for developing and validating advanced neural shape representation models, including implicit neural fields and sparse representations~\citep{Prasad2022}. These models can capture the structural complexity of plants with reduced computational overhead, making them valuable for large-scale phenotyping studies. Researchers can use the dataset to train and evaluate machine learning models that generalize across diverse genotypes and growth conditions, improving the robustness of predictive algorithms~\citep{Jiang2018}.

\paragraph{Virtual Phenotyping and Computational Plant Modeling:}
The dataset supports the exploration of novel methods for 3D shape analysis, including applications in virtual phenotyping and computational plant modeling. For example, the integration of neural representations with segmented point clouds can enable the simulation of plant responses to environmental stimuli, such as light and wind~\citep{marshall2017crops}. These simulations can contribute to optimizing crop designs for improved yield and resource efficiency.

\paragraph{Agricultural Technology Development:}
In addition to scientific research, the dataset has potential applications in agricultural technology development. It can be used to train and validate algorithms for autonomous systems, such as robotic harvesting and precision agriculture tools~\citep{chlingaryan2018}. These systems rely on accurate 3D representations of plants for navigation, crop monitoring, and targeted interventions.

\subsection{Virtual Light Simulation with HELIOS}
\label{subsec:helios_simulation}

We performed a virtual light interception experiment using the HELIOS 3D modeling framework~\citep{bailey2019helios}, to demonstrate a practical application of MaizeField3D. The goal was to evaluate the canopy-level light absorption across different maize varieties reconstructed from the dataset and validate the simulated results against field measurements.

\begin{figure}[b!]
    \centering
    \begin{subfigure}[b]{0.85\linewidth}
        \centering
        \includegraphics[trim=0cm 2cm 0cm 3cm, clip, width=\linewidth]{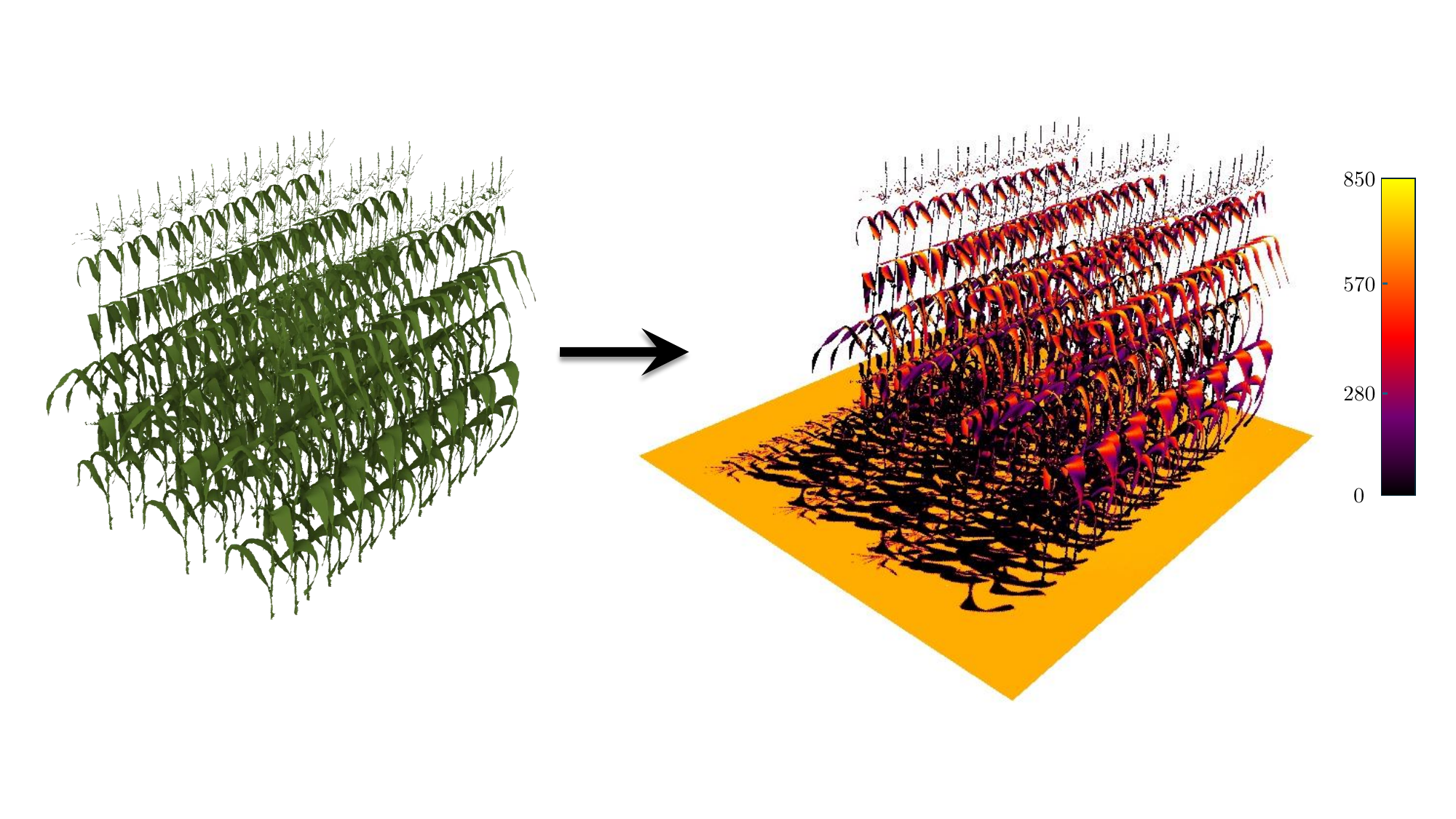}
        \caption{}
        \label{fig:helios_simulation}
    \end{subfigure}
    \begin{subfigure}[b]{0.45\linewidth}
        \centering
        \includegraphics[width=\linewidth]{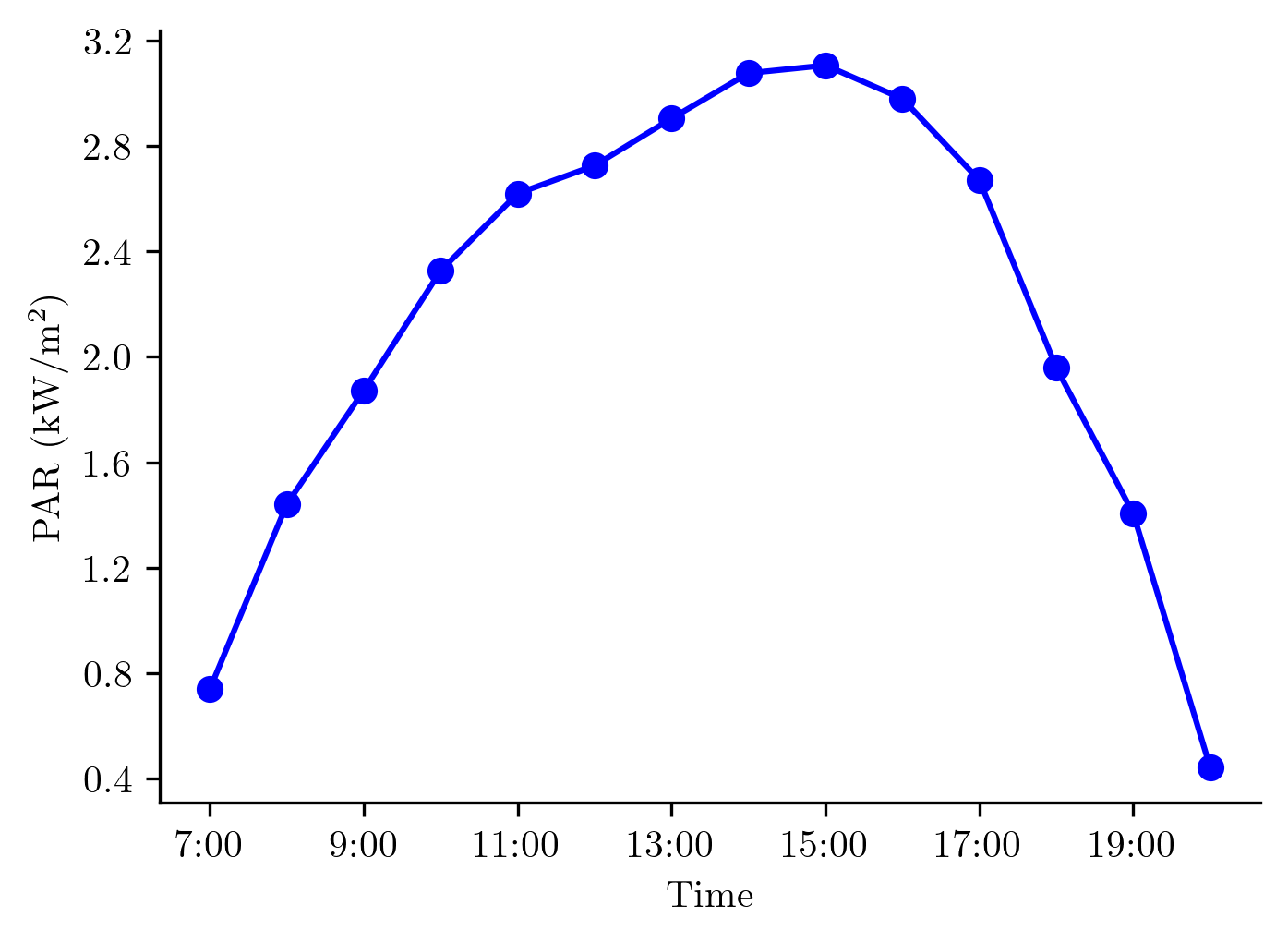}
        \caption{}
        \label{fig:helios_simulation_hour}
    \end{subfigure}
    \caption{Schematic of the HELIOS simulation. (a) A procedurally generated virtual field (left) based on MaizeField3D and the radiation simulation output (right) showing spatially varying absorbed PAR. (b) Hourly intercepted PAR (kW/m$^2$) on the reconstructed field for a typical day.}
\end{figure}

Each 3D plant used in the simulation consisted of procedurally generated NURBS leaf surfaces derived from MaizeField3D, combined with stalk meshes reconstructed using the ball pivoting surface reconstruction algorithm in MeshLab. These reconstructions ensured realistic structural representations for simulation purposes. To generate genotype-specific virtual field plots, we created 42 unique structural variants for each of five selected genotypes. These variants were generated by applying randomized leaf rotations to the base plant, simulating natural morphological variability within a genotype. Specifically, for each variant, we rotated individual leaves around either the local stalk tangent (azimuthal rotation, up to ±10°) or a perpendicular axis (leaf tilt, up to ±5°). These augmented point clouds preserve genotype-level features while introducing realistic intra-genotype diversity~\cite{lee2002expanding}. We then assembled each virtual field plot using these 42 variants in a 3-row by 14-plant grid, following agronomic spacing standards (30 inches between rows, 6 inches between plants)~\citep{saleem2025accessing}. This resulted in five simulated field plots, each corresponding to a different maize plant shown in \figref{fig:procedural_output}.

\figref{fig:helios_simulation} illustrates the visualization of the HELIOS simulation process. On the left, we show a procedurally generated virtual field constructed using MaizeField3D plant. On the right, we depict the HELIOS simulation output at one representative hour, where absorbed PAR is mapped across the canopy using color gradients. This highlights how structural details captured in MaizeField3D can be directly leveraged in physics-based light interaction models to analyze canopy-scale light distribution. 
We then used HELIOS to simulate diurnal PAR exposure over a full day (7:00 to 20:00). \figref{fig:helios_simulation_hour} shows the intercepted PAR across the day for the same reconstructed field. The curve highlights the peak absorption around midday and the decrease toward morning and evening hours.  The simulation was configured with fixed geographical coordinates (Ames, Iowa). Intercepted PAR was computed per hour using area-weighted radiation fluxes on plant surfaces. The corresponding total daily intercepted PAR values of the full field plots generated from the five plants (\figref{fig:procedural_output}) are reported in \tabref{tab:helios_par}, revealing how structural variability influences light capture.

To validate the accuracy of these simulations, we replicated the experimental setup used for field validation by Zhou et al.~\citep{zhou2024genetic}. As described in their work, field measurements were collected on August 7, 2020, 60 days after planting, during anthesis. The fraction of the PAR (fPAR) interception was estimated by subtracting ground-level measurements from above-canopy measurements and normalizing by the latter. In our simulation, we implemented an equivalent virtual sensing protocol by placing a horizontal sensor bar between the center rows of the simulated field. This allowed direct comparison of simulated fPAR values against field-measured values using the same spatial sampling strategy. We then compared the simulated fPAR values from the reconstructed fields against field-measured fPAR values for the same genotypes. As shown in \tabref{tab:helios_par}, the simulated and measured fPAR values show strong agreement. This result supports the accuracy of our simulation pipeline in capturing canopy-scale light interception behavior across different genotypes.

\begin{table}[t!]
\centering
\caption{Total intercepted PAR energy and fractional PAR (fPAR) values for five simulated fields on a typical summer day. The plant names correspond to the labels shown in \figref{fig:procedural_output}.}
\label{tab:helios_par}
    \begin{tabular}{p{0.1\linewidth}>{\raggedleft\arraybackslash}p{0.25\linewidth}>{\raggedleft\arraybackslash}p{0.25\linewidth}>{\raggedleft\arraybackslash}p{0.25\linewidth}}
    \toprule
    \textbf{Plant} & \textbf{Total PAR (kWh/m$^2$)} & \textbf{Model fPAR } & \textbf{Measured fPAR} \\
    \midrule
    a & 42.44 & 0.96 & 0.96 \\
    b & 31.97 & 0.87 & 0.95 \\
    c & 28.26 & 0.86 & 0.76 \\
    d & 30.84 & 0.64 & 0.65 \\
    e & 30.26 & 0.84 & 0.94 \\
    \bottomrule
    \end{tabular}
\end{table}

\section{Conclusions}

This study presents a high-resolution 3D point cloud dataset of maize plants. By capturing detailed structural features and providing comprehensive metadata, this dataset bridges a critical gap in agricultural research, enabling precise and scalable analyses of plant architecture, morphology, and genotype-phenotype relationships. The dataset’s segmentation, annotation, and standardized downsampling cater to a wide range of downstream applications, from feature extraction and shape representation to the development of machine learning models for plant phenotyping. 

While the dataset represents a step forward in providing high-resolution 3D point clouds for plant phenomics, limitations must be acknowledged. A key limitation is the focus on a single growth stage. The dataset captures plants at vegetative maturity, providing valuable insights into their structural characteristics at this stage. However, it does not encompass the full developmental cycle of maize, such as seedling, flowering, or senescence stages. This restricts the scope of analyses to specific phenotypic traits relevant to vegetative maturity. Future work will aim to address these limitations by expanding the dataset and incorporating additional metadata fields to enhance its utility. Specifically, future versions of the dataset will include data from multiple growth stages, enabling researchers to study dynamic phenotypic changes throughout the plant lifecycle. These additions will provide a more comprehensive resource for investigating developmental processes and genotype-phenotype interactions over time. 

We also demonstrated a practical application of MaizeField3D by performing a virtual light interception experiment using the HELIOS 3D modeling framework. This use case illustrates how the dataset’s structurally accurate reconstructions can be leveraged in functional–structural plant modeling, enabling simulation of canopy-level light absorption under realistic field conditions. Such examples highlight the dataset’s utility for hypothesis testing and environmental interaction studies beyond static analysis.

By providing this dataset to the research community, we aim to accelerate progress in plant science and crop improvement studies. The integration of detailed metadata further enhances its usability, allowing researchers to tailor their analyses to specific traits and research objectives.  We encourage researchers to explore this dataset in diverse agricultural contexts, from studying plant responses to environmental conditions to improving crop yield and sustainability. Future expansions of this resource, including additional growth stages and environmental data, will further amplify its impact, fostering innovation and collaboration across the field.

\section*{Data Availability}

The MaizeField3D dataset is publicly available on the Hugging Face Datasets platform at 
\url{https://huggingface.co/datasets/BGLab/MaizeField3D}. It includes high-resolution point clouds, segmented plant models, metadata, and reconstructed outputs in STL and DAT formats. Full documentation and download instructions are provided on the project website: \url{https://baskargroup.github.io/MaizeField3D/}. The code for procedural NURBS surface generation used in this work is available at
\url{https://github.com/baskargroup/ProceduralMaize3D}.

\section*{Acknowledgments}
This work is supported by the AI Research Institutes program [\textit{AI Institute for Resilient Agriculture (AIIRA)}, Award No. 2021-67021-35329] from the National Science Foundation and U.S. Department of Agriculture’s National Institute of Food and Agriculture. We acknowledge support from the Iowa State University Plant Science Institute. YL was supported in part by NSF grants (DBI-1661475 and IOS-1842097) to PSS and others, and a scholarship from the China Scholarship Council. 

\section*{Author CRediT Contributions}
Conceptualization: AB, TZJ, AK, BG; Data curation: EK, MH, YL; Formal Analysis: MH, TZJ; Funding acquisition: AK, PSS, BG; Investigation: EK, MH, JG; Methodology: EK, MH, JG, AB, TZJ, AK, BG; Project administration: AK, PSS, BG; Resources: YL, PSS, BG; Software: EK, MH, JG, TZJ; Supervision: AB, TZJ, AK, PSS, BG; Validation: EK, MH, YL; Visualization: EK, MH; Writing – original draft: EK, MH, TZJ, AK, BG; Writing – review \& editing: MH, TZJ, YL, AK, PSS, BG
\newpage

\bibliography{references.bib}

\clearpage
\renewcommand{\thefigure}{S.\arabic{figure}}
\setcounter{figure}{0}
\renewcommand{\thetable}{S\arabic{table}}
\setcounter{table}{0}

\section*{Supplementary Materials}
\label{sec:Supplementary}

\begin{figure}[H]
    \centering

    \begin{subfigure}[b]{0.3\textwidth}
        \includegraphics[trim=10in 0in 10in 0in, clip, width=\textwidth]{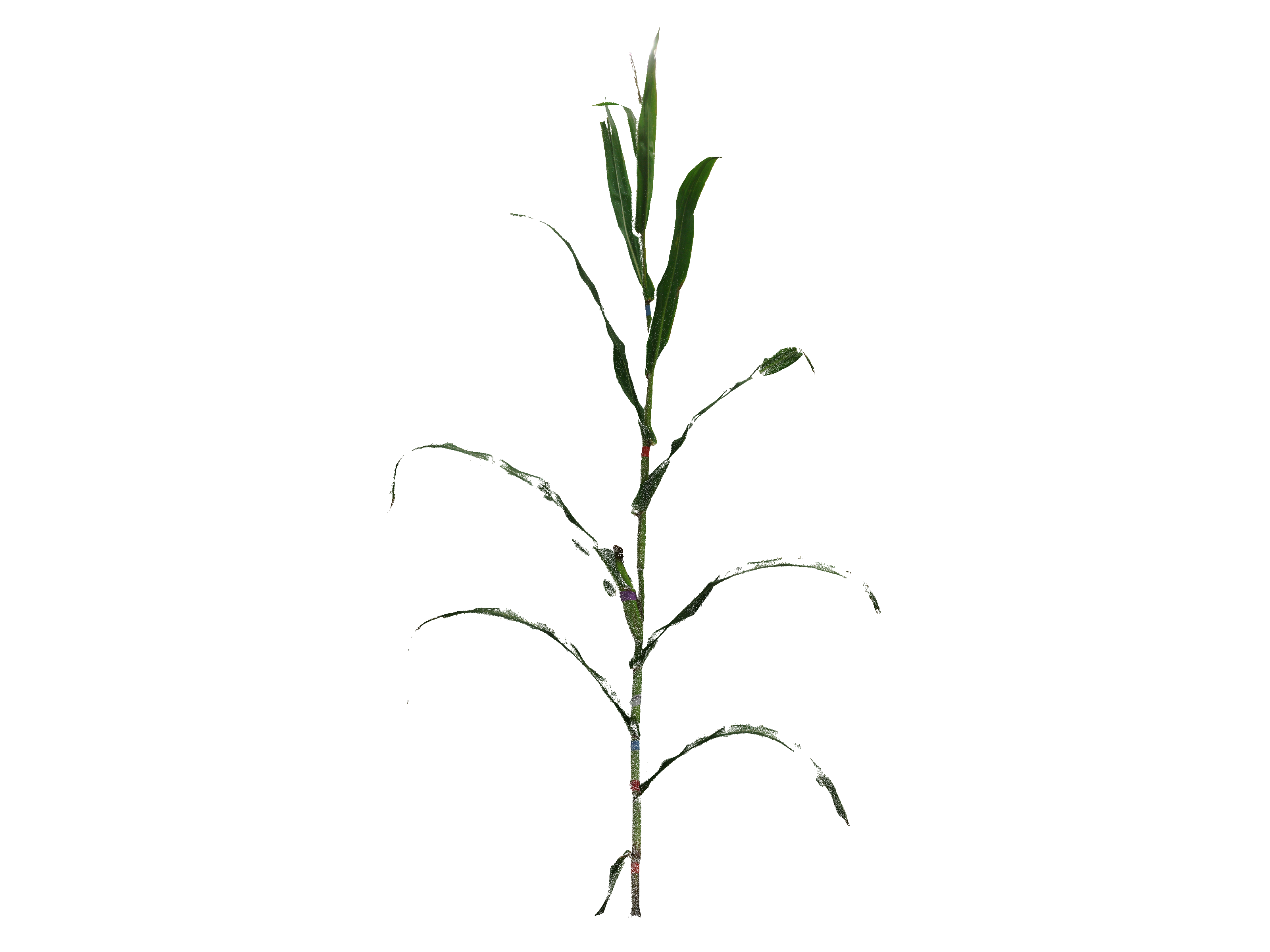}
    \end{subfigure}
    \begin{subfigure}[b]{0.3\textwidth}
        \includegraphics[trim=10in 0in 10in 0in, clip, width=\textwidth]{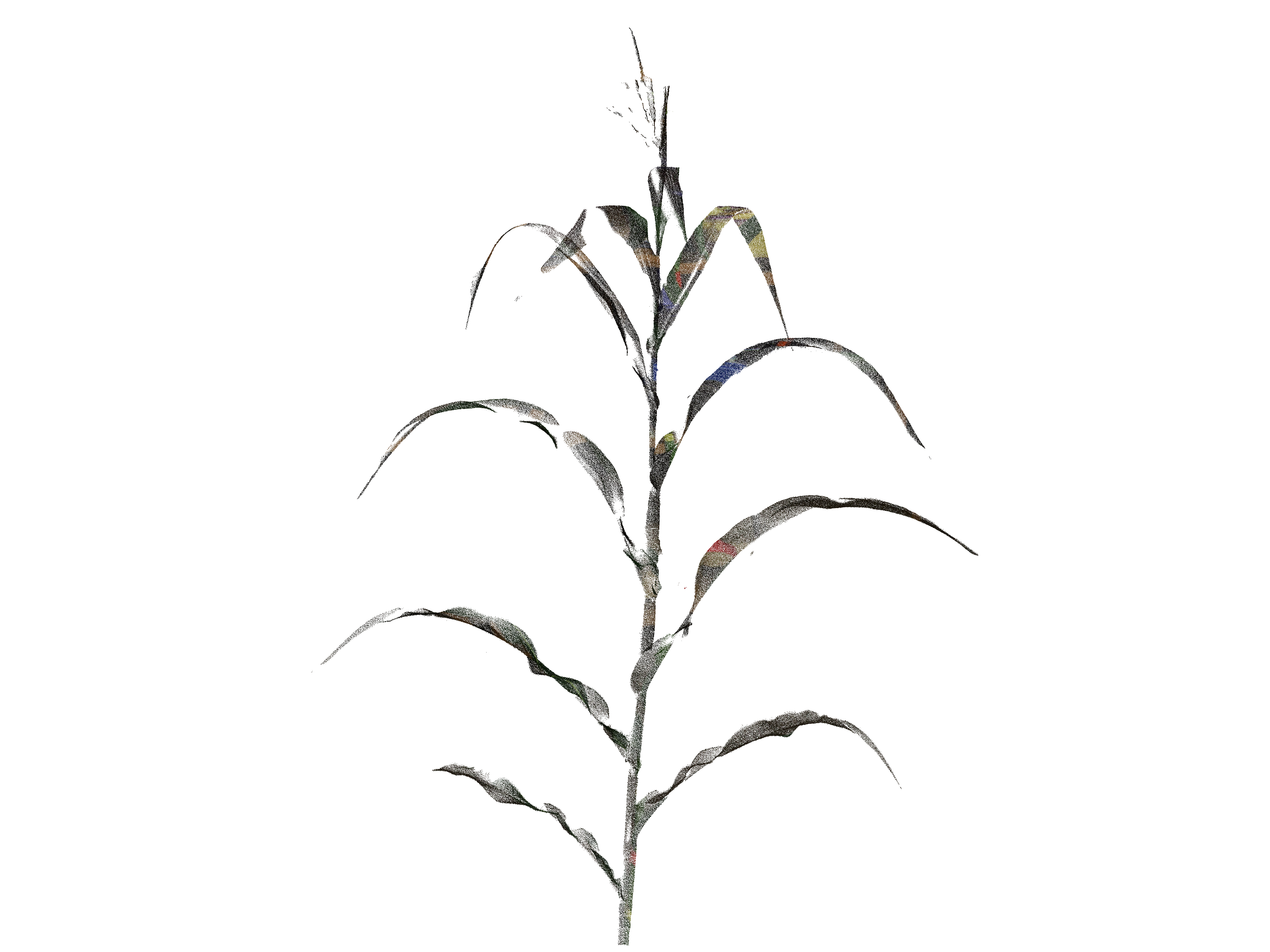}
    \end{subfigure}
    \begin{subfigure}[b]{0.3\textwidth}
        \includegraphics[trim=10in 0in 10in 0in, clip, width=\textwidth]{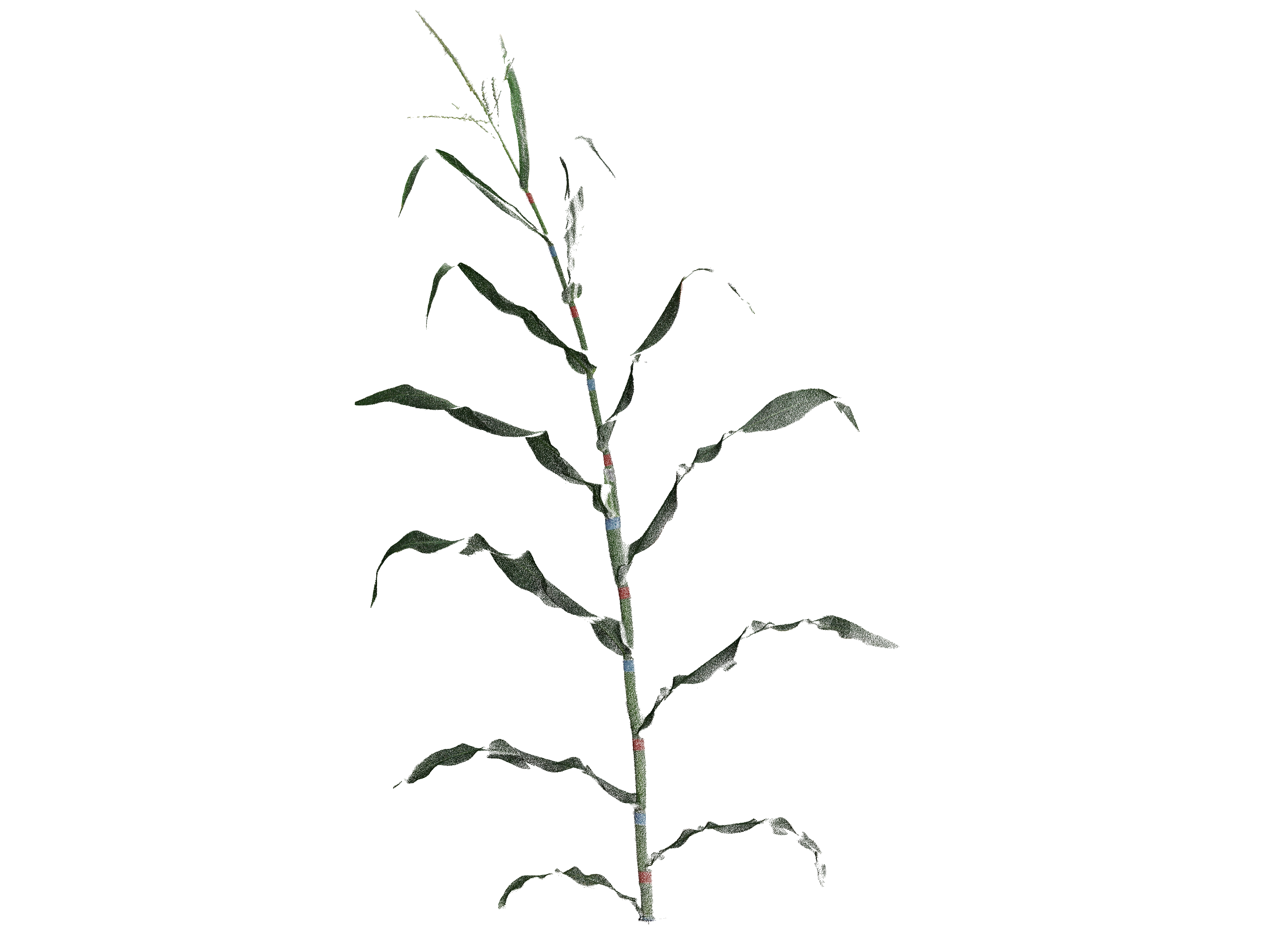}
    \end{subfigure}

    \begin{subfigure}[b]{0.3\textwidth}
        \includegraphics[trim=9in 0in 10in 0in, clip, width=\textwidth]{Figures_M/21-JN3761-1.png}
    \end{subfigure}
    \begin{subfigure}[b]{0.3\textwidth}
        \includegraphics[trim=10in 0in 10in 0in, clip, width=\textwidth]{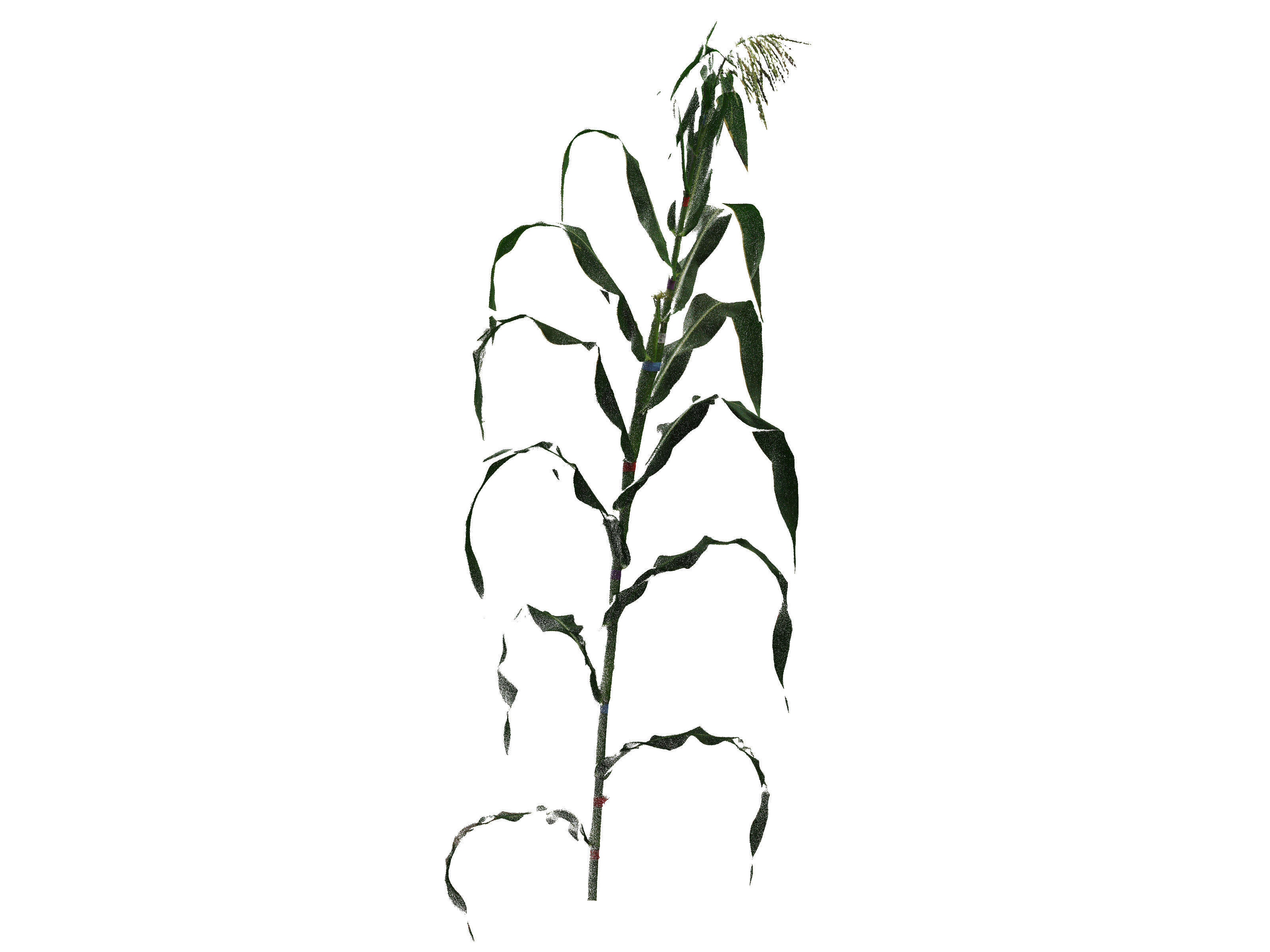}
    \end{subfigure}
    \begin{subfigure}[b]{0.3\textwidth}
        \includegraphics[trim=10in 0in 10in 0in, clip, width=\textwidth]{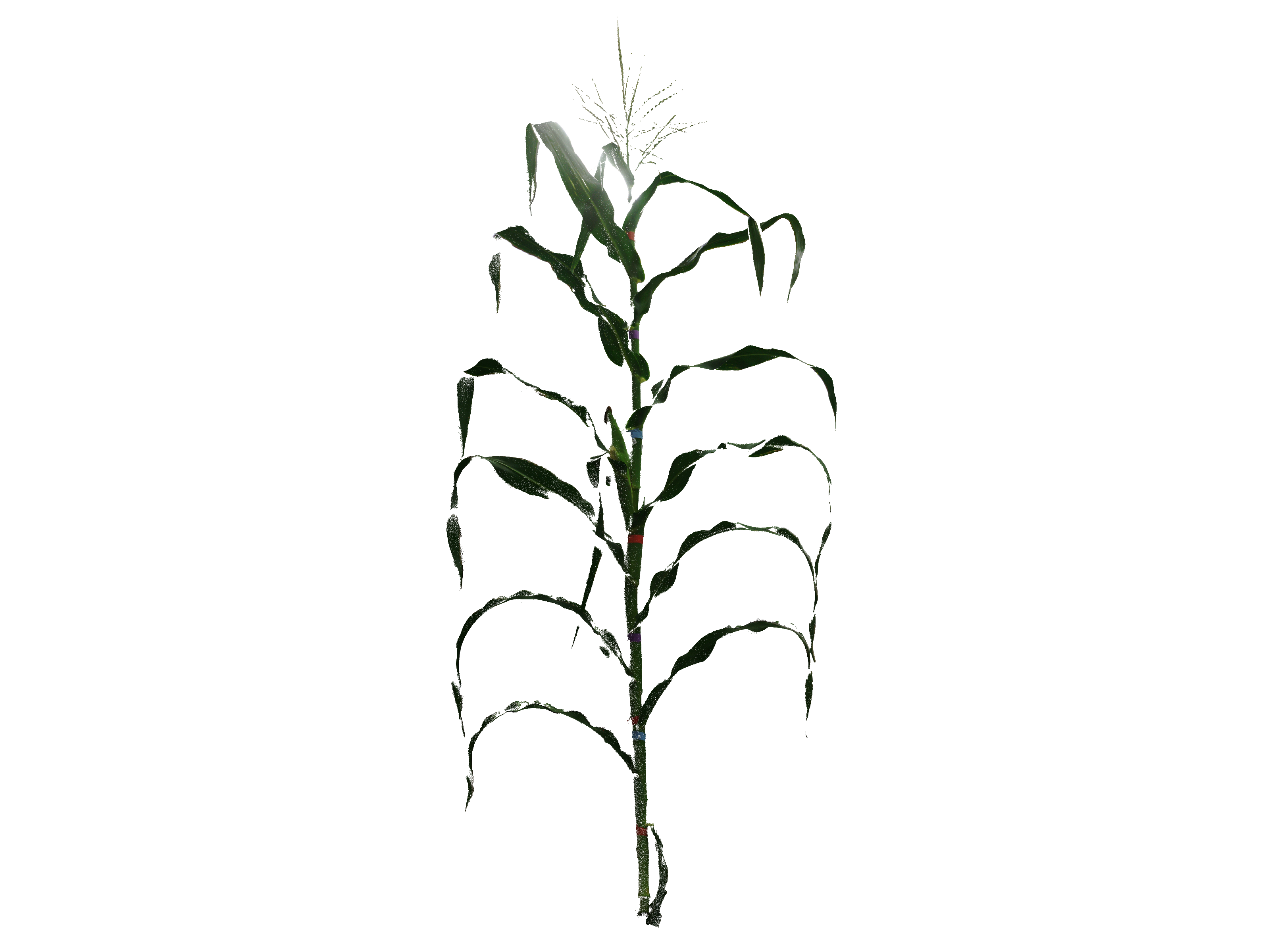}
    \end{subfigure}

    \caption{Additional maize plant point clouds from the MaizeField3D dataset, showcasing the original data. These examples further illustrate the morphological diversity observed across maize plant varieties.}
    \label{fig:additional_original_plants}
\end{figure}

\begin{figure}[t!]
    \centering
    \begin{subfigure}[b]{0.3\textwidth}
        \includegraphics[trim=10in 0in 10in 0in, clip, width=\textwidth]{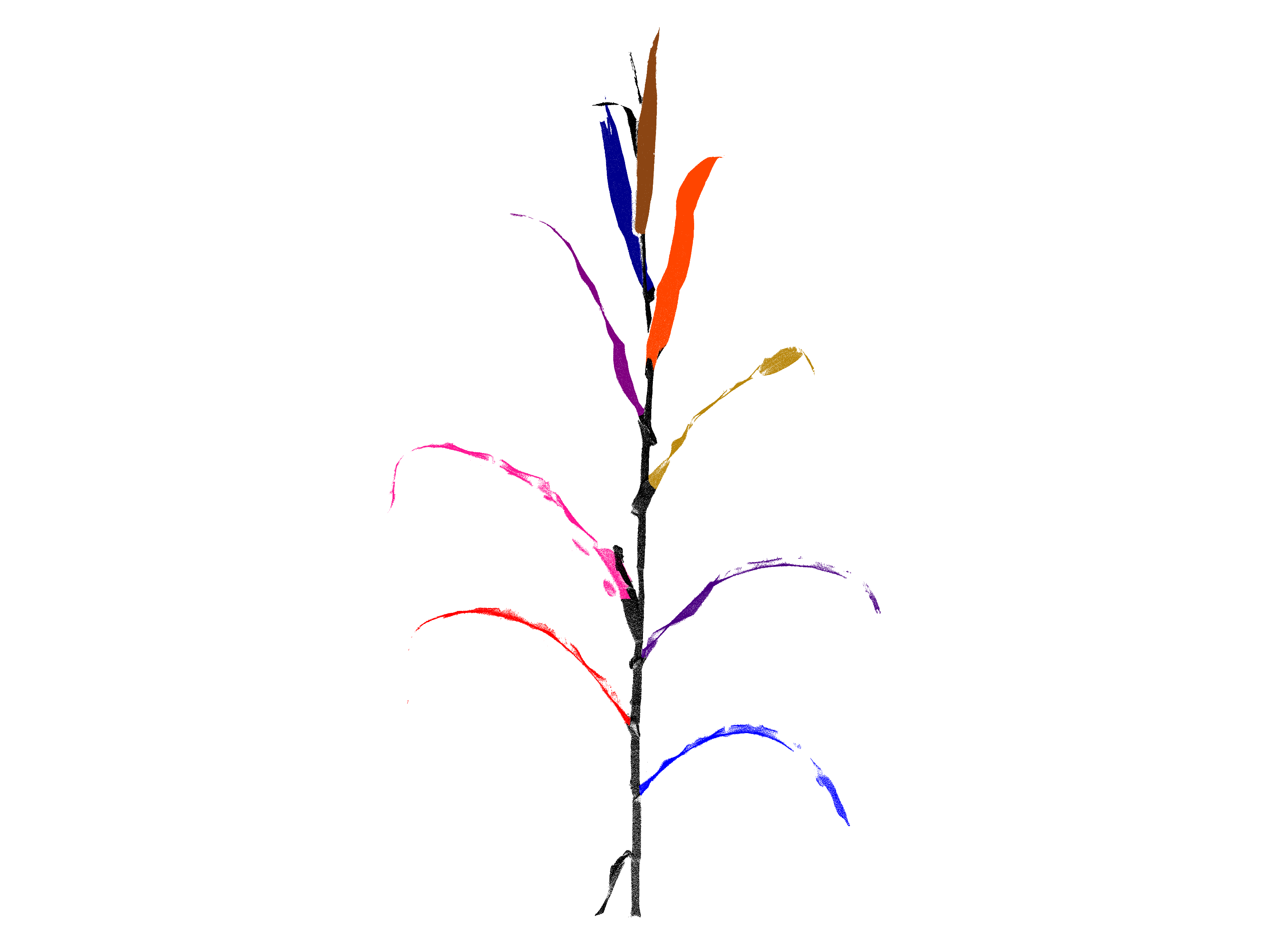}
    \end{subfigure}
    \begin{subfigure}[b]{0.3\textwidth}
        \includegraphics[trim=10in 0in 10in 0in, clip, width=\textwidth]{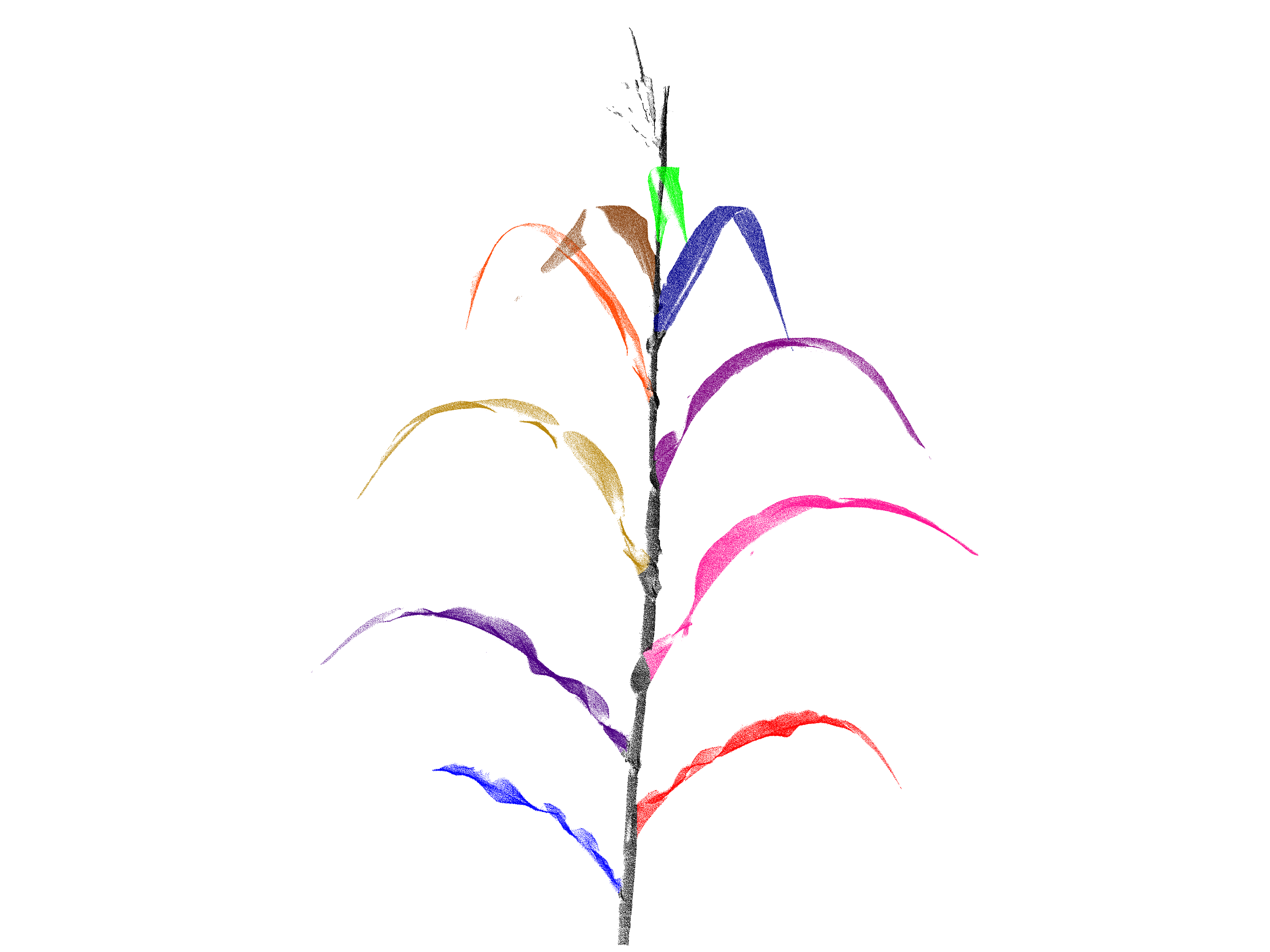}
    \end{subfigure}
    \begin{subfigure}[b]{0.3\textwidth}
        \includegraphics[trim=10in 0in 10in 0in, clip, width=\textwidth]{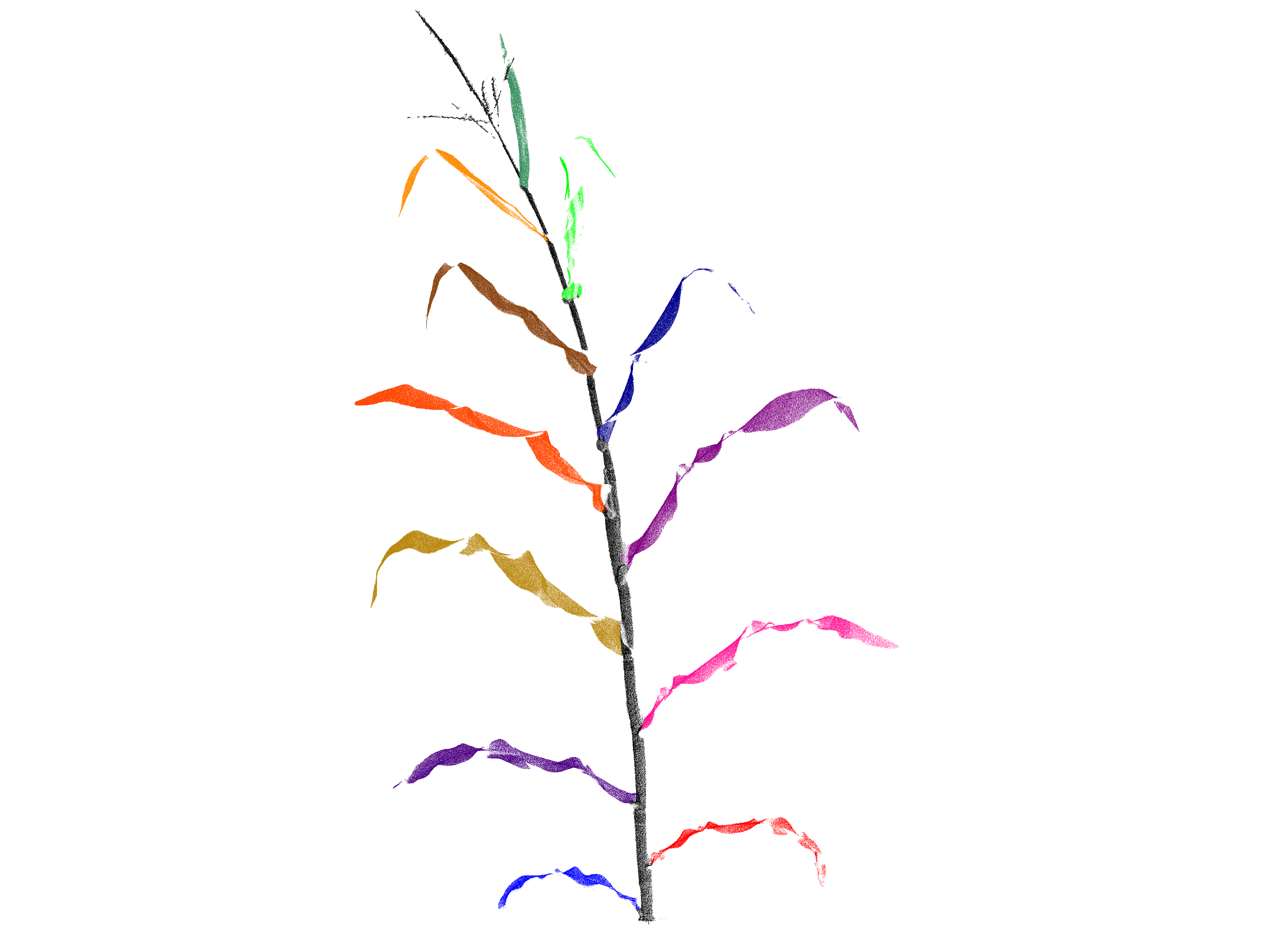}
    \end{subfigure}
    \vspace{0.1in}
    \begin{subfigure}[b]{0.3\textwidth}
        \includegraphics[trim=9in 0in 10in 0in, clip, width=\textwidth]{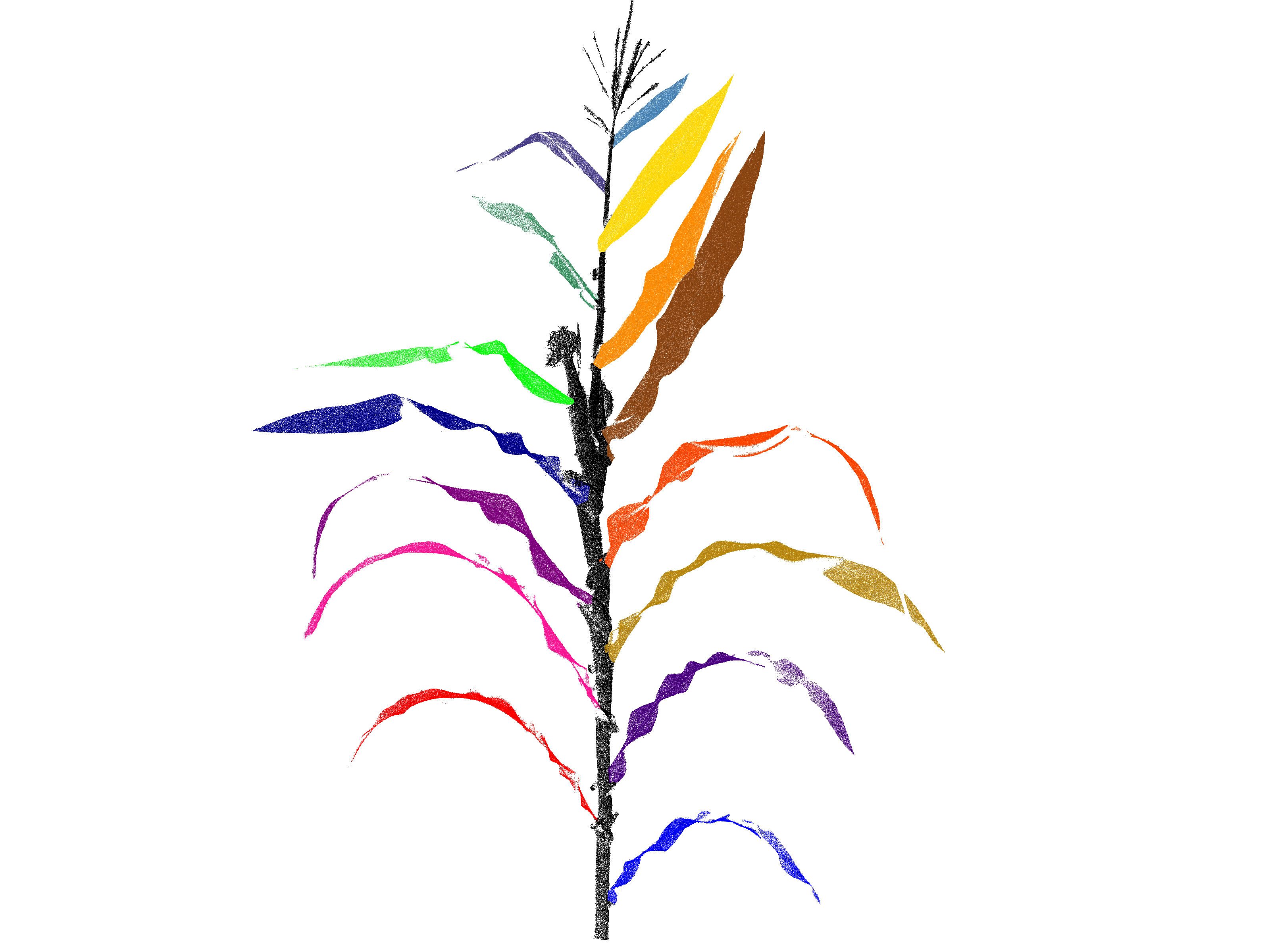}
    \end{subfigure}
    \begin{subfigure}[b]{0.3\textwidth}
        \includegraphics[trim=10in 0in 10in 0in, clip, width=\textwidth]{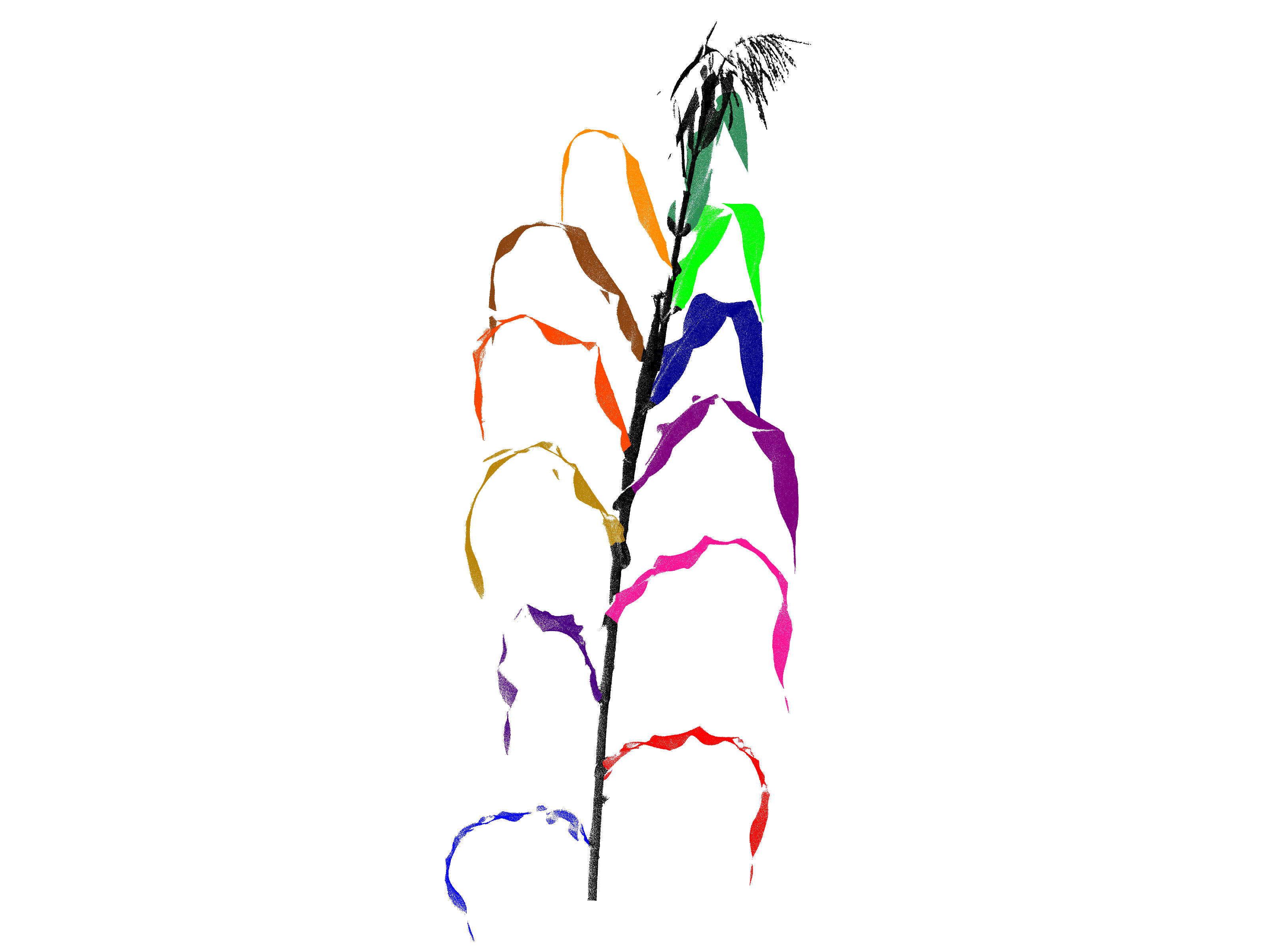}
    \end{subfigure}
    \begin{subfigure}[b]{0.3\textwidth}
        \includegraphics[trim=10in 0in 10in 0in, clip, width=\textwidth]{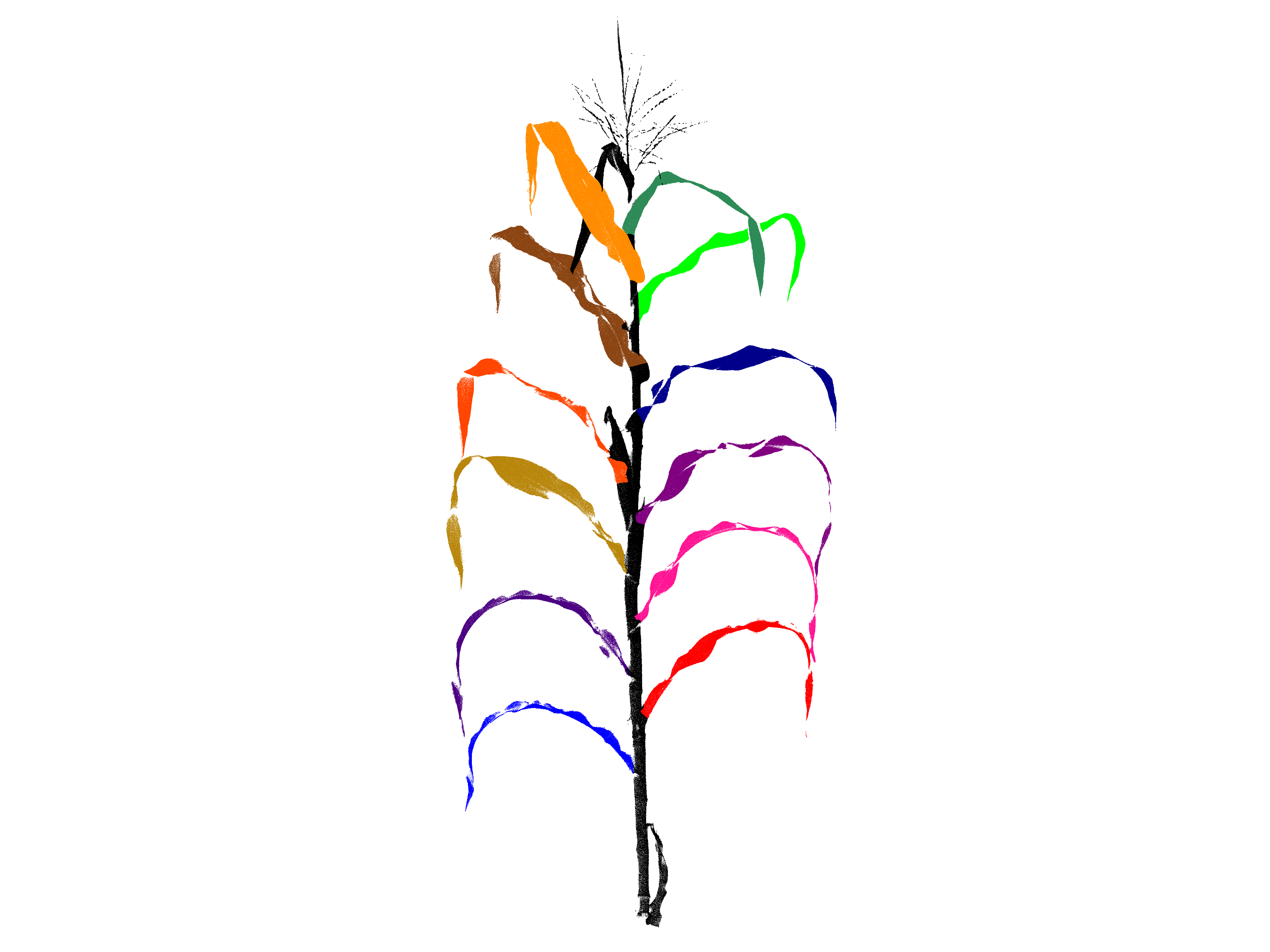}
    \end{subfigure}

    \caption{Additional examples of segmented maize plant point clouds from the dataset.}
    \label{fig:additional_segmented_plants}
\end{figure}

\begin{table}[h!]
\centering
\caption{Color Coding for Plant Components}
\label{tab:color_coding}
\begin{tabular}{|c|c|c|c|}
\hline
\textbf{Component}  & \textbf{Color Name}   & \textbf{RGB Values} & \textbf{Hex Code} \\ \hline
Leaf 1              & Bright Blue           & [0, 0, 255]         & \#0000FF          \\ \hline
Leaf 2              & Bright Red            & [255, 0, 0]         & \#FF0000          \\ \hline
Leaf 3              & Indigo                & [75, 0, 130]        & \#4B0082          \\ \hline
Leaf 4              & Deep Pink             & [255, 20, 147]      & \#FF1493          \\ \hline
Leaf 5              & Dark Goldenrod        & [184, 134, 11]      & \#B8860B          \\ \hline
Leaf 6              & Purple                & [128, 0, 128]       & \#800080          \\ \hline
Leaf 7              & Orange Red            & [255, 69, 0]        & \#FF4500          \\ \hline
Leaf 8              & Dark Blue             & [0, 0, 139]         & \#00008B          \\ \hline
Leaf 9              & Saddle Brown          & [139, 69, 19]       & \#8B4513          \\ \hline
Leaf 10             & Bright Green          & [0, 255, 0]         & \#00FF00          \\ \hline
Leaf 11             & Dark Orange           & [255, 140, 0]       & \#FF8C00          \\ \hline
Leaf 12             & Sea Green             & [46, 139, 87]       & \#2E8B57          \\ \hline
Leaf 13             & Gold                  & [255, 215, 0]       & \#FFD700          \\ \hline
Leaf 14             & Dark Slate Blue       & [72, 61, 139]       & \#483D8B          \\ \hline
Leaf 15             & Steel Blue            & [70, 130, 180]      & \#4682B4          \\ \hline
Leaf 16             & Brown                 & [165, 42, 42]       & \#A52A2A          \\ \hline
Stalk               & Black                 & [0, 0, 0]           & \#000000          \\ \hline
\end{tabular}
\end{table}

\begin{table}[h]
\centering
\caption{Runtime comparison (in seconds) of different downsampling strategies at three target sizes.}
\label{tab:downsampling_comparison}
\begin{tabular}{lccc}
\toprule
\textbf{Method} & \textbf{100k} & \textbf{50k} & \textbf{10k} \\
\midrule
Random  & 0.03 ± 0.01 & 0.03 ± 0.01 & 0.02 ± 0.01 \\
Voxel   & 0.95 ± 0.26 & 1.01 ± 0.32 & 1.11 ± 0.34 \\
FPS     & 107.57 ± 38.24 & 55.99 ± 19.59 & 11.26 ± 3.93 \\
Poisson & 108.30 ± 37.22 & 57.01 ± 19.95 & 11.58 ± 4.09 \\
\bottomrule
\end{tabular}
\end{table}

\end{document}